\pdfoutput=1
\documentclass[11pt]{article}

\usepackage[preprint]{acl}

\usepackage{times}
\usepackage{latexsym}
\usepackage{xspace}
\usepackage{tabularx}
\usepackage{adjustbox}
\usepackage[T1]{fontenc}

\usepackage[utf8]{inputenc}

\usepackage{microtype}

\usepackage{inconsolata}

\usepackage{graphicx}

\usepackage{amsmath,amssymb}
\usepackage{booktabs}
\usepackage{multirow}
\usepackage{booktabs}
\usepackage{graphicx}
\usepackage{hyperref}
\usepackage{enumitem}
\usepackage{tikz}
\usepackage{subcaption}

\newcommand{\postspace}{\vskip -3mm}
\newcommand{\minipostspace}{\vskip -2mm}

\newcommand{\method}{AutoPyVerifier\xspace}

\definecolor{cadmiumgreen}{rgb}{0.0, 0.42, 0.24}
\definecolor{cardinal}{rgb}{0.77, 0.12, 0.23}
\definecolor{cadmiumred}{rgb}{0.89, 0.0, 0.13}

\usepackage[most]{tcolorbox}
\newtcolorbox[list inside=prompt,auto counter,number within=section]{prompt}[1][]{
    fontupper=\ttfamily\footnotesize,
    boxsep=5pt,
    left=0pt,
    right=0pt,
    top=0pt,
    bottom=0pt,
    boxrule=1pt,
    breakable,
    #1,
}

\usepackage{paracol}
\usepackage{caption}

\usepackage{xcolor}
\usepackage{listings}
\usepackage[most]{tcolorbox}
\tcbuselibrary{listings,skins,breakable}

\definecolor{codebg}{RGB}{31,31,31}
\definecolor{codeborder}{RGB}{70,70,70}
\definecolor{codetext}{RGB}{220,220,220}
\definecolor{codekw}{RGB}{86,156,214}      
\definecolor{funcgreen}{RGB}{78,201,176}   
\definecolor{codecom}{RGB}{106,153,85}
\definecolor{codestr}{RGB}{206,145,120}
\definecolor{codenum}{RGB}{140,140,140}

\lstdefinestyle{mypython}{
    language=Python,
    alsoletter={_},
    basicstyle=\fontsize{6}{7}\ttfamily\color{codetext},
    identifierstyle=\color{codetext},
    keywordstyle=\color{codekw}\bfseries,
    commentstyle=\color{codecom},
    stringstyle=\color{codestr},
    numberstyle=\fontsize{5}{6}\selectfont\color{codenum},
    numbers=left,
    numbersep=6pt,
    showstringspaces=false,
    breaklines=false, 
    columns=fixed, 
    breakatwhitespace=false,
    keepspaces=true,
    tabsize=2,
    frame=none,
    emph={
        _text,
        _normalize,
        _extract_final,
        _num,
        _candidate_claims,
        reference_supported_final_answer,
        no_multiple_distinct_final_claims,
        no_obvious_unfinished_or_repetitive_reasoning,
        _num_from_string,
        final_answer_present_and_parseable,
        internal_numeric_consistency,
        _normalize,
        matches_reference_answer,
        aggregate,
        get_words,
        verify_instruction_constraints,
        check_no_hallucinated_success,
        check_no_clarification_or_refusal,
        check_expected_functions_called,
        ends_with_assistant_text,
        no_suspicious_uncertainty_comments,
        constraint_aware_bruteforce_filter,
        small_oracle_or_sample_signal,
        no_obvious_malformed_or_truncated_code,
        task_contract_compatible,
        python_parseable_basic,
        structural_task_match,
        code_response_present,
        _clean,
        _txt,
        _large_problem
    },
    emphstyle=\color{funcgreen}\bfseries
}

\newtcblisting{pyblock}[2][]{
    enhanced,
    colback=codebg,
    colframe=codeborder,
    boxrule=0.5pt,
    arc=2pt,
    left=3pt,
    right=3pt,
    top=3pt,
    bottom=3pt,
    listing only,
    listing engine=listings,
    listing options={style=mypython},
    title={#2},
    coltitle=white,
    fonttitle=\bfseries,
    #1
}

\title{\method: Learning Compact Executable Verifiers for Large Language Model Outputs}

\author{Pouya Pezeshkpour\\
Megagon Labs\\
\texttt{pouya@megagon.ai} \\
\And
Estevam Hruschka\\
Megagon Labs \\
\texttt{estevam@megagon.ai}}

\begin{document}
\maketitle
\begin{abstract}
Verification is becoming central to both reinforcement-learning-based training and inference-time control of large language models (LLMs). Yet current verifiers face a fundamental trade-off: LLM-based verifiers are expressive but hard to control and prone to error, while deterministic executable verifiers are reliable and interpretable but often limited in capability. We study the following question: given a development set of LLM outputs and labels for a target objective, such as correctness, can we automatically induce a \emph{minimal} set of Python verifiers whose joint satisfaction closely matches that objective?  
We propose \textbf{\method}, a framework that uses an LLM to synthesize candidate verifier functions and then refines them through search over a directed acyclic graph (DAG).   
By navigating the DAG, \method systematically explores the space of deterministic executable verifiers and selects a compact verifier set whose joint satisfaction best approximates the target objective. 
Across mathematical reasoning, coding, function calling, and instruction-following benchmarks for several state-of-the-art LLMs, \method improves target-objective prediction by up to \textbf{55.0} F1 points over the initial LLM-generated verifier sets. Additional analyses show that the most useful verification targets vary by benchmark and model, and that the DAG-based search shifts the learned verifier sets toward more structural and semantically grounded checks. We further show that exposing the discovered verifier set to an LLM as an external tool improves downstream accuracy by up to \textbf{17.0} points. We release our code\footnote{\url{https://github.com/megagonlabs/AutoPyVerifier}}.

\end{abstract}

\section{Introduction}
Recent progress in large language models (LLMs) has improved performance on mathematical reasoning, code generation, and many other complex tasks \cite{comanici2025gemini, singh2025openai}. At the same time, verification has become a core component of modern LLM systems. In reinforcement-learning-based training, verifiers provide supervision by determining which trajectories should be rewarded, filtered, or refined \cite{lightman2023let, guo2025deepseek}. At inference time, verification supports reranking, search, pruning, and self-correction \cite{madaan2023self, snell2024scaling}. It is also becoming central to reflection-based and agentic frameworks, where models evaluate intermediate outputs, incorporate feedback, and improve subsequent actions \cite{yao2022react, ma2025advancing, zhang2026verified}. As a result, the quality of a reasoning system increasingly depends not only on its ability to generate outputs, but also on its ability to \emph{verify} them reliably. 

\begin{figure}[t!]
    \centering
    \includegraphics[width=\linewidth]{figs/overview.png}
\caption{\textbf{\method overview:} Given task inputs and description, LLM outputs, and objective labels, \method initializes candidate Python verifier sets and iteratively refines them through DAG-based search guided by a utility function, returning a compact verifier set aligned with the target objective.}
    \label{fig:overview}
    \postspace
\end{figure}

Current verification methods face a basic trade-off. LLM-based verifiers and judges are flexible and often powerful, but they can be hard to control, inconsistent, and overly sensitive to surface plausibility rather than true objective alignment \cite{gu2024survey, wang2025trustjudge, li2025llms}. Deterministic verifiers, such as Python functions, are interpretable, executable, and often highly accurate when applicable, but they are usually narrow in scope and costly to design for each task \cite{huangbeyond,ficek2025scoring}. In practice, neither extreme is ideal: LLM verifiers may be unreliable, while programmatic ones often lack the coverage needed for realistic reasoning tasks.

This trade-off suggests a different direction: rather than treating LLM-based and deterministic verification as fixed alternatives, we ask whether deterministic verifiers can be \emph{induced} from data to better match a desired objective. Specifically, the question we aim to answer is the following: given LLM outputs on a development set and labels for a target objective---such as correctness, validity, or successful task completion---can we automatically construct a \emph{small} set of Python verifiers whose joint satisfaction closely matches that objective? Such a verifier set offers several practical benefits: it is more controllable and interpretable than an LLM judge, more accurate and comprehensive than task-specific one-off checks, and can be used not only for evaluation, but also as actionable feedback for downstream inference, reranking, or repair, as well as a reward signal in reinforcement-learning-based training.

We propose \textbf{\method} (Figure \ref{fig:overview}), a framework for automatically constructing compact Pyhton verifier sets from labeled model outputs. 
Given a task, a development set, and a target objective, \method first uses an LLM to synthesize several initial Python verifier sets, which are inserted as children of a root node in a DAG. Each node in this graph corresponds to a candidate set of verifiers. \method then iteratively grows the DAG by selecting a node for expansion and producing a small number of modified children, for example by adding a new verifier function or revising an existing one. 
The choice of which node to expand is guided by an objective that combines predictive quality (such as F1 against the target labels), exploration value, verifier-set size, and feasibility. 
This search procedure enables \method to efficiently explore the space of executable verifier sets and ultimately select a small, interpretable set whose joint satisfaction best approximates the desired objective. 
Our motivation is to treat verification as a first-class optimization problem rather than a fixed manually engineered component. Instead of assuming the correct verification logic is known, \method aims to \emph{discover} it from model behavior and objective labels.

We evaluate \method on coding, mathematical reasoning, function calling, and instruction-following benchmarks using several state-of-the-art LLMs. Our experiments address three main questions: whether compact induced verifier sets can accurately approximate the target objective, whether they generalize across base models, and whether they remain useful when exposed back to an LLM as inference-time tools. We find that \method consistently improves over the initial LLM-generated verifier sets, with gains of up to \textbf{55.0} F1 points, while keeping the learned verifier sets compact at only 1-6 functions. Many of the learned verifiers also transfer effectively to outputs from a different base model, suggesting that they capture reusable task-level verification logic rather than merely model-specific patterns. Our ablations further show that the importance of exploration, compactness, and feasibility depends on both the benchmark and the model, and that search shifts the learned verifier sets toward more structural and semantically grounded checks. Finally, when exposed as external tools during inference, the discovered verifier sets improve downstream accuracy by up to \textbf{17.0} percentage points.

\begin{figure*}[t!]
    \centering
    \includegraphics[width=0.8\linewidth]{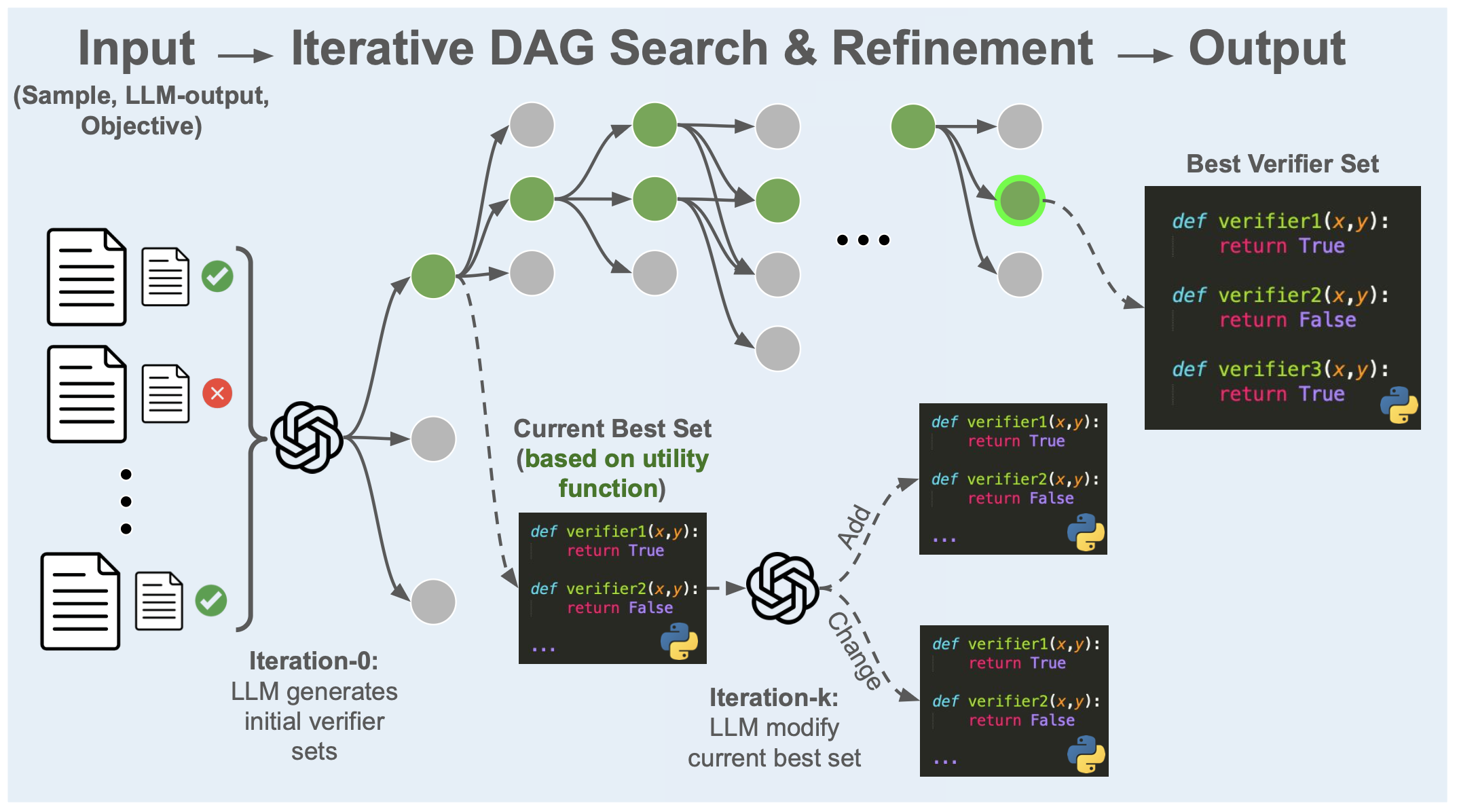}
\caption{Given labeled examples consisting of task inputs (together with the task description), LLM outputs, and objective labels, \method first prompts an LLM to generate initial Python verifier sets. These sets are inserted as nodes in a DAG and iteratively refined by selecting the current best node under the utility function and expanding it through LLM-proposed modifications. The search returns the best verifier set found, yielding a compact set of executable Python verifiers aligned with the target objective.}
    \label{fig:method}
    \postspace
\end{figure*}
\section{\method}
\label{sec:method}
\method is a framework for inducing a small set of executable Python verifiers whose joint behavior closely matches a target objective defined over model outputs. 
Rather than assuming that the correct verification logic is known in advance, \method treats verifier construction as a search problem over candidate verifier sets. 
At a high level, \method starts from several LLM-generated verifier sets, organizes them as nodes in a DAG, and then iteratively expands the graph by modifying promising nodes. Figure~\ref{fig:method} presents an overview of the full pipeline.

\subsection{Problem Setup}

Let
\[
\mathcal{D} = \{(x_i, y_i, \ell_i)\}_{i=1}^{N}
\]
denote a development set, where \(x_i\) is a task input, \(y_i\) is a model output for that input, and \(\ell_i\) is the label for a target objective defined on the behavior of the model output. $\ell_i$ can represent a binary, multi-label, or even free-form text objective. 

\paragraph{Verifier functions.}
We consider a verifier as an executable Python function
\[
v : \mathcal{X} \times \mathcal{Y} \times \mathcal{Z} \rightarrow \{0,1\},
\]
which takes a task input \(x\) and a model output \(y\), required extracted values from the model output \(z\), and returns whether a particular property is satisfied. A verifier set is denoted by
\[
V = \{v_1, v_2, \dots, v_m\}.
\]
Given a verifier set \(V\), we denote by \(\hat{\ell}_V\) the induced prediction of that set on the development examples through a fixed aggregation rule over the verifier outputs.
We focus on learning a \emph{set} of verifiers rather than a single comprehensive verifier. A single verifier that fully captures the target objective would be harder to induce and less interpretable for both humans and LLMs. A verifier set instead decomposes the objective into smaller executable checks, making the resulting verification logic easier to interpret, analyze, and use downstream.

\paragraph{Objective.}
Our goal is to construct a \emph{compact} verifier set \(V^\star\) such that \(\hat{\ell}_{V^\star}\) aligns as closely as possible with the target labels \(\ell\), while keeping the number of verifier functions small. Formally, \method searches for a verifier set that optimizes predictive quality with respect to the target objective under an explicit complexity penalty:
\[
V^\star = \arg\max_{V \in \mathcal{V}} \; \text{Utility}(V),
\]
where \(\mathcal{V}\) is the search space of verifier sets.

\subsection{Initialization: LLM Synthesis and DAG Construction}

\paragraph{Initial verifier synthesis.}
Given a task description, the development set \(\mathcal{D}\), and the target objective, \method first prompts an LLM to generate several initial sets of Python verifiers. Each generated set is intended to capture some aspect of the target objective, such as structural validity, consistency, formatting constraints, execution behavior, or task-specific correctness signals.

\paragraph{DAG representation.}
\method organizes the search space as a directed acyclic graph
\[
G = (\mathcal{N}, \mathcal{E}),
\]
with a root node \(r\). Each child of the root corresponds to one initial LLM-generated verifier set. More generally, each node \(n \in \mathcal{N}\) represents a candidate verifier set \(V(n)\), and each directed edge \((n, n') \in \mathcal{E}\) corresponds to a transformation that produces a new verifier set from an existing one. 
This representation provides a natural way to track the search history, preserve parent-child relations, support iterative refinement of verifier sets, and make the search process more efficient.

\subsection{DAG Search: Scoring, Selection, and Expansion}

\paragraph{Node scoring.}
At each iteration, \method assigns an acquisition value to each candidate node:
\begin{align*}
\mathrm{acq}(n) =\;& \mathrm{TaskScore}(n) \\
&+ \alpha \cdot \mathrm{ExplorationScore}(n) \\
&- \beta \cdot |V(n)| \\
&+ \gamma \cdot \mathrm{FeasibilityScore}(n).
\end{align*}
where \(|V(n)|\) is the number of verifier functions in node \(n\), and \(\alpha,\beta,\gamma\) are task-dependent hyperparameters. 
For binary objectives, we define
\[
\mathrm{TaskScore}(n) = \mathrm{F1}(n),
\]
that is, the F1 score between the node's induced predictions and the target labels on the development set. For multi-label or free-form objectives, \(\mathrm{TaskScore}(n)\) can be replaced by a metric aligned with the corresponding task objective.

\paragraph{Exploration.}
To avoid focusing only on currently high-scoring nodes and to account for future potential, \method includes a UCB-inspired exploration bonus:
\[
\mathrm{ExplorationScore}(n)
=
\sqrt{
\frac{\log(1 + T)}{1 + t(n)}
},
\]
where \(T\) is the total number of node visits so far, and \(t(n)\) is the number of times node \(n\) has been selected for expansion. This bonus favors under-explored nodes while decaying as a node is visited repeatedly.

\paragraph{Compactness and feasibility.}
Because we seek a \emph{minimal} verifier set, \method penalizes nodes with larger numbers of functions through the term \(-\beta \cdot |V(n)|\). This discourages unnecessarily large verifier sets, which can increase overhead and risk of overfitting. 
In addition, \method uses a binary feasibility signal:
\[
\mathrm{FeasibilityScore}(n)=
\begin{cases}
1, & \parbox[t]{0.62\columnwidth}{if $\mathrm{TP}(n) > 0.5$\\ and $\mathrm{TN}(n) > 0.5$,} \\[6pt]
0, & \text{otherwise.}
\end{cases}
\]
where \(\mathrm{TP}(n)\) and \(\mathrm{TN}(n)\) denote the ratios of correctly identified positive and negative examples, respectively. This term favors nodes that are non-degenerate on both sides of the label space.

\paragraph{Node selection and critic-guided child generation.}
At each iteration, \method selects the node with the highest acquisition value for expansion. Let \(n^\star\) denote the selected node, with verifier set \(V(n^\star)\). To refine this node, \method first computes its error profile on the development set, including the false-positive set
\[
\mathcal{E}_{\mathrm{FP}}(n^\star) = \{(x_i,y_i,\ell_i) \in \mathcal{D} : \hat{\ell}_{V(n^\star),i}=1, \ell_i=0\},
\]
and the false-negative set
\[
\mathcal{E}_{\mathrm{FN}}(n^\star) = \{(x_i,y_i,\ell_i) \in \mathcal{D} : \hat{\ell}_{V(n^\star),i}=0, \ell_i=1\}.
\]
\method then makes a first LLM call to a \emph{critic}, which receives the task description, the current verifiers code, summary statistics of the node, and representative false-positive and false-negative examples (chosen randomly). The critic returns a structured diagnosis of the current verifier set, identifying: (i) loopholes responsible for false positives, (ii) over-restrictions responsible for false negatives, (iii) redundant or weak checks, (iv) missing verifier types or missing context fields, and (v) suggested edits such as \texttt{ADD}, \texttt{REMOVE}, \texttt{REPLACE}, or \texttt{MODIFY}.

In a second LLM call, \method invokes a \emph{modifier}, conditioned on the current verifier bundle, the critic summary, and the same false-positive and false-negative examples. The modifier produces a small set of children
\[
\mathrm{Child}(n^\star) = \{n_1', \dots, n_K'\},
\]
where each child corresponds to a revised verifier set obtained by editing \(V(n^\star)\). Typical edits include adding a new verifier, replacing or modifying an existing weak verifier, removing redundant checks, changing the aggregation logic, or introducing stronger context requirements through updated \texttt{requires} fields. The modifier is explicitly encouraged to keep bundles compact and deterministic, and to prefer improving weak checks rather than only increasing the number of verifiers. Each generated child is then evaluated on the development set, inserted into the DAG, and becomes eligible for future expansion. 
\subsection{Final Selection and Output}

\paragraph{Search loop.}
\method repeats the cycle of scoring, node selection, child generation, and evaluation until a stopping condition is met, such as a search budget or convergence of the best node.

\paragraph{Returned verifier set.}
At the end of the search, \method returns the final verifier set
\[
V^\star = V(\hat{n}),
\]
where \(\hat{n}\) is selected by ranking all explored nodes first according to their task score and then, among nodes with comparable score, according to verifier-set size.

\paragraph{Usage.}
The returned verifier set can be used in several ways: as an executable proxy for the target objective, as a reranking or filtering signal, as structured feedback for downstream repair, or as a reward component in RL-based training.

\begin{table*}[t!]
\small
\centering
\begin{tabular}{clrrrrrrrrr}
\toprule 
&\multirow{2}{*}{\bf Model} & \multicolumn{2}{c}{\bf AIME}&\multicolumn{2}{c}{\bf LiveCodeBench}&\multicolumn{2}{c}{\bf ComplexFuncBench}&\multicolumn{2}{c}{\bf IFBench}\\
\cmidrule(lr){3-4}
\cmidrule(lr){5-6}
\cmidrule(lr){7-8}
\cmidrule(lr){9-10}
&&Initial&Final&Initial&Final&Initial&Final&Initial&Final\\
\midrule
\multirow{5}{*}{\rotatebox{90}{ID}}&GPT-5.4&38.8& \bf 80.1 \textcolor{green!50!black}{(+41.3)}&\bf 47.4&\bf 79.6 \textcolor{green!50!black}{(+32.2)}&30.1&85.1 \textcolor{green!50!black}{(+55.0)}&28.5&42.6 \textcolor{green!50!black}{(+14.1)}\\
&GPT-5.4 mini&\bf 56.6&73.8 \textcolor{green!50!black}{(+17.2)}&43.7&56.8 \textcolor{green!50!black}{(+13.1)}&27.7&62.4 \textcolor{green!50!black}{(+34.7)}&25.7&39.5 \textcolor{green!50!black}{(+13.8)}\\
&Gemini-3.1 Pro&46.7&62.5 \textcolor{green!50!black}{(+15.8)}&38.6&64.0 \textcolor{green!50!black}{(+25.4)}&\bf 55.6&\bf 96.4 \textcolor{green!50!black}{(+40.8)}&29.4&\bf 81.5 \textcolor{green!50!black}{(+52.1)}\\
&Gemini-3.1 Fl&38.8&51.5 \textcolor{green!50!black}{(+12.7)}&37.8&49.9 \textcolor{green!50!black}{(+12.1)}&47.9&84.9 \textcolor{green!50!black}{(+37.0)}&\bf 41.1&51.7 \textcolor{green!50!black}{(+10.6)}\\
&Claude Hai-4.5&38.8&73.1 \textcolor{green!50!black}{(+34.3)}&37.8&37.8 \textcolor{white!50!black}{(0.00)}&27.7&54.0 \textcolor{green!50!black}{(+26.3)}&39.5&59.1 \textcolor{green!50!black}{(+19.6)}\\
\midrule
\multirow{5}{*}{\rotatebox{90}{OOD}}&GPT-5.4&50.1&74.8 \textcolor{green!50!black}{(+24.7)}&43.9&\bf 78.7 \textcolor{green!50!black}{(+34.8)}&\bf 49.7&49.7 \textcolor{white!50!black}{(0.00)}&25.1&37.1 \textcolor{green!50!black}{(+12.0)}\\
&GPT-5.4 mini&\bf 69.1&73.3 \textcolor{green!50!black}{(+4.20)}&\bf 47.7&53.9 \textcolor{green!50!black}{(+6.20)}&\bf 49.7&46.1 \textcolor{red!50!black}{(-3.6)}&22.4&41.5 \textcolor{green!50!black}{(+19.1)}\\
&Gemini-3.1 Pro&66.1&77.3 \textcolor{green!50!black}{(+11.2)}&41.2&58.8 \textcolor{green!50!black}{(+17.6)}&23.7&\bf 74.3 \textcolor{green!50!black}{(+50.6)}&25.1&\bf 79.5 \textcolor{green!50!black}{(+54.4)}\\
&Gemini-3.1 Fl&41.2&\bf 91.6 \textcolor{green!50!black}{(+50.4)}&41.2&58.8 \textcolor{green!50!black}{(+17.6)}&\bf 49.7&55.3 \textcolor{green!50!black}{(+5.60)}&33.5&41.3 \textcolor{green!50!black}{(+7.80)}\\
&Claude Hai-4.5&34.1&53.3 \textcolor{green!50!black}{(+19.2)}&41.2& 41.2 \textcolor{white!50!black}{(0.00)}&\bf 49.7& 49.7 \textcolor{white!50!black}{(0.00)}&\bf 41.5&63.1 \textcolor{green!50!black}{(+21.6)}\\
\bottomrule
\end{tabular}
\caption{F1 performance of \method-induced verifier sets in predicting whether model outputs satisfy the task objective in both in-distribution (ID, GPT-4.1) and out-of-distribution (OOD, Gemini-2.5 Flash-Lite) settings.}
\label{tab:main_res}
\postspace
\end{table*}

\section{Experimental Details}
\label{sec:exp-detail}
\paragraph{Benchmarks.}
We evaluate \method on benchmarks chosen to stress verification on reasoning tasks. In particular, we prioritize settings that satisfy three properties: (i) evaluation is hidden or externally executed, reducing the chance of directly optimizing for the final answer; (ii) multiple candidate solutions can be generated for each task instance, making reranking and verifier-based selection meaningful; and (iii) there is room for structural verification beyond surface-level correctness. Following these criteria, we include AIME (2024 and 2025) \cite{balunovic2025matharena} for mathematical reasoning, LiveCodeBench \cite{jain2024livecodebench} using problems released after 2025 to reduce contamination concerns, and ComplexFuncBench \cite{zhong2025complexfuncbench}, where we focus on the cross-domain, multi-step, and constrained samples to create a more challenging setting for verifier induction. We also include IFBench \cite{pyatkin2025generalizing} as an additional benchmark covering instruction follwing capability of LLMs.

\paragraph{Models.}
For in-distribution candidate generation, we use GPT-4.1 as the backbone model. To study out-of-distribution generalization, we additionally evaluate on Gemini 2.5 Flash-Lite \cite{comanici2025gemini}. For consistency, each full \method pipeline uses the same underlying LLM across all of its components, including initial verifier synthesis, critic generation, and verifier refinement. We consider five state-of-the-art LLMs as the engine behind \method: GPT-5.4, GPT-5.4 mini \cite{singh2025openai}, Gemini-3.1 Pro, Gemini-3.1 Flash \cite{comanici2025gemini}, and Claude Haiku 4.5. 
The prompts used in \method, including for verifier generation, critique, refinement, context extraction, and tool use, are provided in Appendix.

\paragraph{Implementation details.}
At initialization and during refinement, \method generates up to three candidate child verifier sets for each expanded node. We run the DAG search for 20 expansion steps. The final verifier set is selected from all explored nodes based primarily on task score, which in our binary setting is F1, with verifier-set size used as a secondary preference when scores are tied. We tune the hyperparameters of acquisition function (\(\alpha\), \(\beta\), and \(\gamma\)) using grid search for each task (more details in Appendix).

\section{Experiments}
In this section, we study five questions: whether \method can learn compact verifier sets that predict the target objective, whether those verifiers transfer across base models, which utility term is most impactful,  what kinds of verification logic \method discovers, and whether the learned verifiers can improve downstream LLM performance when used as tools.

\begin{figure*}[t!]
    \centering
    \includegraphics[width=\linewidth]{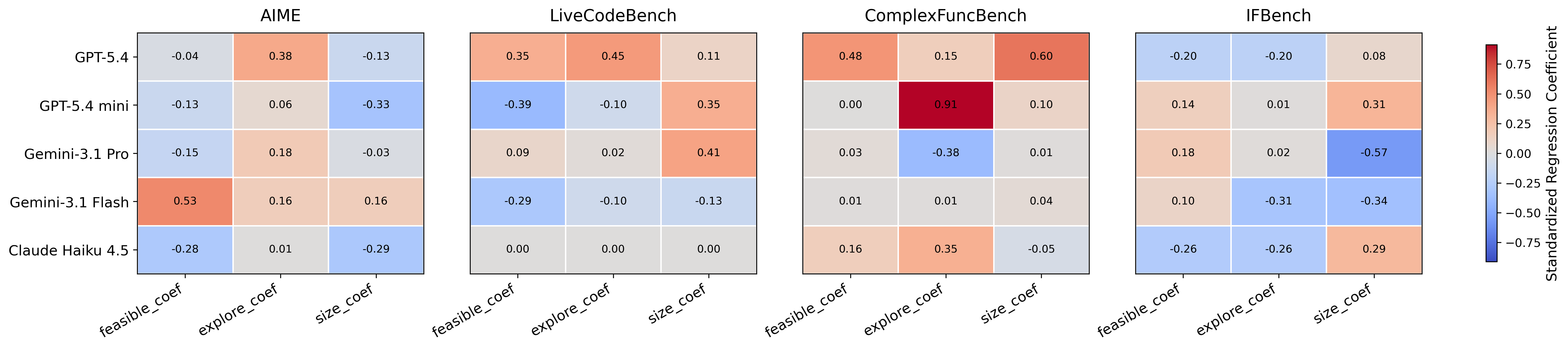}
    \caption{Linear regression coefficients showing the effect of \method objective terms on final verifier performance.}
    \label{fig:heatmap}
    \postspace
\end{figure*}
\subsection{In-Distribution Verifier Learning}
\label{sec:ID}

We begin with an in-distribution setting where GPT-4.1 generates candidate outputs for each benchmark instance, and each output receives a binary label from the task evaluator (e.g., on AIME the objective is final-answer correctness).  Using this labeled development set, we run \method with each underlying model, tune \(\alpha\), \(\beta\), and \(\gamma\), and report the best run performance of both the \emph{initial} verifier set from the first synthesis step and the \emph{final} verifier set selected after search. All numbers in Table~\ref{tab:main_res} are F1 scores. In every case, the learned verifier sets remain compact, containing only 1-6 verifier functions. We provide both the initial and final verifier sets for the best-performing model in Appendix.

Table~\ref{tab:main_res} shows that \textbf{the search stage consistently strengthens the verifier sets over their initial LLM-generated versions across almost all in-distribution settings}, with gains of up to \textbf{55.0 F1}. The improvements are especially pronounced on ComplexFuncBench and IFBench, where first step synthesis often yields relatively weak initial verifier sets but search is able to refine them substantially. 
Across the table, stronger frontier models such as GPT-5.4 and Gemini-3.1 Pro often achieve some of the largest gains after search, while smaller or lighter models still benefit substantially, indicating that the refinement stage is broadly useful rather than tied to a single model family. At the same time, \textbf{the magnitude of improvement varies across models and benchmarks}, suggesting that the usefulness of search depends both on the quality of the initial verifier proposals and on how much room there is for refinement. 

\subsection{Out-of-Distribution Transfer}

We next evaluate whether the verifier sets learned in the in-distribution setting transfer to outputs from a different base model. Concretely, we use Gemini-2.5 Flash-Lite to generate answers for each benchmark instance, obtain the corresponding binary objective labels from the benchmark evaluator, and then apply exactly the same \emph{initial} and \emph{final} verifier sets from the previous subsection, without any re-tuning or additional search. Our goal here is to test whether the learned verifiers capture task-level objective structure rather than merely overfitting to GPT-4.1-specific output patterns.

As shown in the OOD portion of Table~\ref{tab:main_res}, the learned verifier sets often transfer well to outputs generated by a different base model, with improvements of up to \textbf{54.4 F1}. While the OOD results are naturally more variable than the in-distribution ones, the overall pattern still suggests that \textbf{\method frequently learns verification logic that generalizes beyond the model used during induction}. Stronger LLMs such as GPT-5.4 and Gemini-3.1 Pro often result in high quality verifiers after transfer, but smaller models can show substantial performance, indicating that the induced verifier sets are not tied to a single model family.  
Moreover, some verifier sets transfer only weakly, and a small number of settings show no gain or even a slight drop, such as GPT-5.4 and GPT-5.4 mini on ComplexFuncBench. Overall, however, the pattern suggests that \method often learns verification logic that generalizes beyond the model used during induction. This is encouraging because it indicates that the discovered verifier sets are not merely fitting surface regularities of GPT-4.1 outputs, but can in many cases act as reusable executable proxies for the underlying task objective across different model families.

\begin{figure*}[t!]
    \centering
    \includegraphics[width=\linewidth]{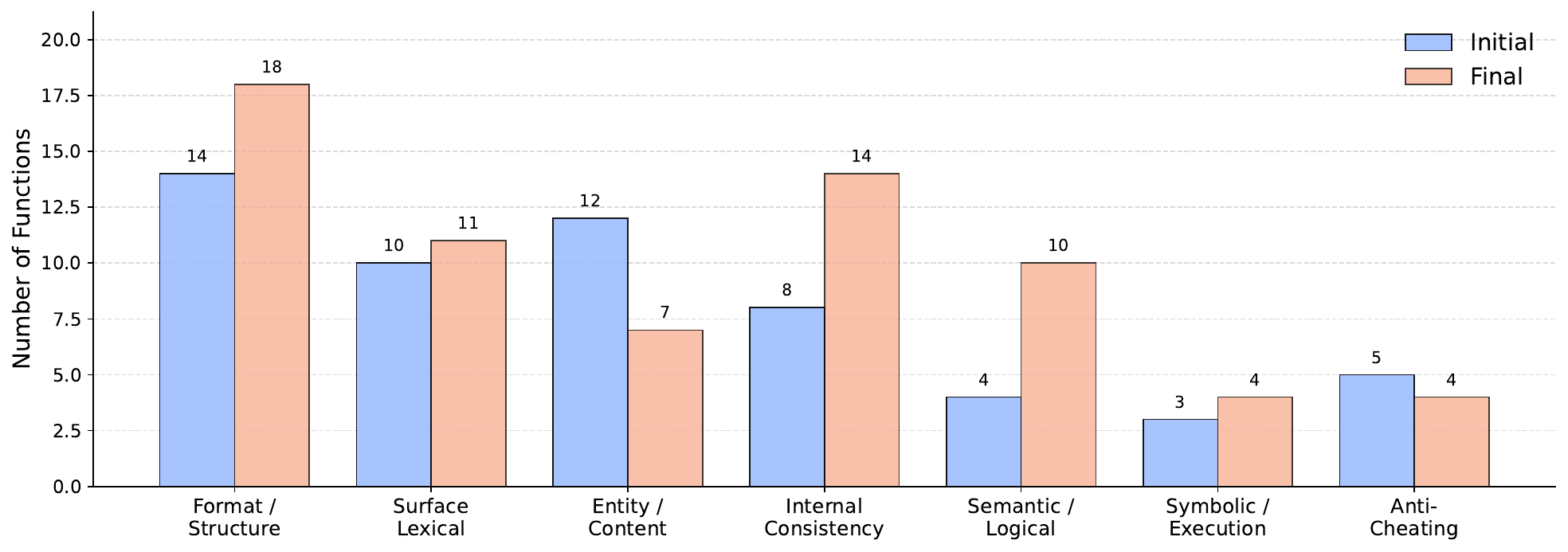}
    \caption{Shift in the distribution of verifier categories before and after applying \method.}
    \label{fig:taxonomy}
    \minipostspace
\end{figure*}
\subsection{Which \method Objective Terms Matter?}

To understand which parts of the \method acquisition objective matter most in practice, we analyze how the search coefficients affect the quality of the learned verifier sets. The acquisition function includes task score, exploration, size penalty, and feasibility, but since task score is always present and not tuned, we focus on the three hyperparameters \(\alpha\), \(\beta\), and \(\gamma\). For each LLM--benchmark pair, we use \method runs (Section \ref{sec:ID}) over a grid of hyperparameter values, record the final best verifier F1, and fit a linear regression model that predicts the F1 from the hyperparameters. Figure~\ref{fig:heatmap} shows the standardized regression coefficients, where positive values indicate that increasing the corresponding term helps the final verifier, and negative values indicate the opposite.

Figure~\ref{fig:heatmap} shows that the relative importance of \method objective terms varies substantially across benchmarks and models. On AIME, exploration is the most consistently helpful term, while stronger size penalties also often help by favoring smaller verifier bundles. In contrast, LiveCodeBench more often benefits from allowing somewhat larger verifier sets, and the effect of exploration is mixed. ComplexFuncBench and IFBench exhibit the greatest heterogeneity, with the same term sometimes helping one model but hurting another. Overall, these patterns suggest that \textbf{there is no single universally optimal weighting of the \method objective terms}, and that \textbf{the best balance between exploration, compactness, and feasibility depends meaningfully on both the benchmark and the underlying LLM}. This supports the use of task-specific hyperparameter tuning and highlights \method's ability to adapt its search behavior across verification regimes.

\subsection{Quality of Learned Verifiers}

To characterize what kinds of verification logic \method discovers, we analyze the \emph{content} of the learned verifier functions rather than only their aggregate predictive performance. Following recent works \citep{wen2024benchmarking, zhang2025cfbench, yang2025structeval, lee2025checkeval, pyatkin2025generalizing}, we define seven verifier categories: \emph{Format / structure}, \emph{Surface lexical}, \emph{Entity / content presence}, \emph{Internal consistency}, \emph{Semantic / logical correctness proxy}, \emph{Lightweight symbolic / execution / numeric}, and \emph{Anti-cheating / leakage control} (more details in Appendix). For each LLM-benchmark pair, we take the initial and final verifier sets from the best hyperparameter run, use GPT-5.4 to assign each verifier function to one of these categories, and then aggregate the assignments across all tasks and models to obtain the distribution shown in Figure~\ref{fig:taxonomy}. Upon manual inspection, we find that GPT-5.4 correctly categorizes about 90\% of the verifier functions.

Figure~\ref{fig:taxonomy} reveals a clear shift in the types of verifiers that survive the \method search process. The initial sets contain many \emph{Entity / content presence} checks, but after search their number decrease, while \emph{Internal consistency} and \emph{Semantic / logical correctness proxy} verifier functions increase. This suggests that \textbf{\method tends to move away from shallow presence-based signals and toward more structurally meaningful and semantically grounded checks}. At the same time, \emph{Format / structure} remains the most common category and grows further, indicating that \textbf{structural validity remains an important and broadly useful signal across benchmarks}. Other categories change only modestly. Overall, the distribution shift suggests that \method does not merely accumulate more rules, but refines verifier bundles toward higher-value verification logic.

\begin{table}[t!]
\small
\centering
\begin{tabular}{lrr}
\toprule 
Model&Base&+ Verifiers\\
\midrule
AIME&36.7&43.3 \textcolor{green!50!black}{(+6.6)}\\
LiveCodeBench&39.0& 42.9 \textcolor{green!50!black}{(+3.9)}\\
ComplexFuncBench&61.7&63.3 \textcolor{green!50!black}{(+1.6)}\\
IFBench&34.7&51.7 \textcolor{green!50!black}{(+17.0)}\\
\bottomrule
\end{tabular}
\caption{Downstream task accuracy of GPT-4.1 before and after adding the learned verifier functions as tools.}
\label{tab:tool_use}
\minipostspace
\end{table}

\subsection{Can Learned Verifiers Improve LLM Performance?}

Finally, we study whether the verifier sets induced by \method are useful not only for objective prediction, but also for improving downstream model performance at inference time. For each benchmark, we take the best-performing verifier set from Section \ref{sec:ID} and expose its verifier functions as tools to GPT-4.1. We then keep the original system prompt largely unchanged, adding only a short instruction encouraging the model to call at least one verifier on a draft solution before producing its final answer (detailed prompt is provided in Appendix).

Table~\ref{tab:tool_use} shows that adding the learned verifier sets as tools consistently improves GPT-4.1 across all four benchmarks, with gains of up to \textbf{17.0} percentage point in accuracy. This suggests that \textbf{the learned verifiers provide actionable signals that the base model can use to revise or confirm candidate outputs}. At the same time, the magnitude of improvement varies meaningfully across benchmarks. IFBench shows the largest gain, likely because verifier feedback is relatively direct and easier for the model to act on, whereas ComplexFuncBench shows a smaller improvement, likely due to the task's higher complexity and the larger overall tool set the model must reason over. Overall, these results indicate that learned verifiers can improve inference-time behavior, but their impact depends not only on verifier quality, but also on task difficulty, tool complexity, and the model's ability to use verifier feedback effectively.

\section{Related Work}

Verification is becoming a core ingredient for making LLM systems more reliable, controllable, and adaptive. Yet current approaches still face a trade-off between flexibility and precision, and often lack compact, reusable verification mechanisms that align closely with task objectives.

\paragraph{Verifiers in LLMs.}
A large body of work uses learned verifiers to score final outputs or intermediate reasoning traces, including LLM-as-a-judge systems, process reward models, and more agentic verifier frameworks used during training or inference \cite{gu2024survey, khalifa2025process, pronesti2026beyond, zhang2026agentv}. At the same time, a growing line of work has shown that these learned approaches can suffer from inconsistency, bias, weak grounding, and correlated failure modes between the generator and verifier \cite{wang2025trustjudge,lee2025correctly}. Complementary work instead uses deterministic or externally grounded checks, such as unit tests, execution feedback, theorem provers, and formal validators \cite{pei2025fover, ficek2025scoring}. While these methods provide stronger grounding and interpretability, they are often task-specific, limited in scope, or costly to build and maintain \cite{pei2025fover, ficek2025scoring}.

\paragraph{Iterative refinement.}
A related line of work improves LLM outputs through iterative cycles of generation, critique, and revision, using self-feedback, tool feedback, or external execution signals \citep{kamoi2024can}. More recent self-improving agent frameworks extend this idea beyond single outputs to evolving prompts, contexts, memories, or reusable skills \citep{zhang2025agentic, zhang2026memskill}. 
\emph{AutoHarness} \citep{lou2026autoharness} is perhaps the closest to our setting, as it improves an agent by automatically synthesizing a code harness; related work similarly augments LLMs with executable scaffolding or harness-like control structures  \citep{pan2026natural}.
Our problem is different, however. Rather than synthesizing an environment harness or a single monolithic judge, we induce a compact \emph{set} of executable Python verifiers from labeled model outputs and optimize that set to approximate an explicit target objective.

\section{Conclusion}
We present \method, a verifier-pipeline framework built on a simple idea: verification should be treated as a searchable computational object rather than a fixed post-hoc component. By generating task-specific root verifiers, refining them through search, and optimizing predictive quality, exploration, verifier-set size, and feasibility, \method effectively learns reliable verifiers. Empirically, we find that search consistently improves over the initial LLM-generated verifier sets, often by a large margin, that many learned verifiers transfer across base models, and that exposing the discovered verifier sets back to the model as tools can further improve downstream performance. Our analyses also suggest that \method refines shallow checks into more structural and semantic verification logic.

More broadly, these results point to verifier induction as a promising direction for building more reliable LLM systems. Several natural extensions follow from this perspective: \method could be applied beyond single-model outputs to multi-agent systems, extended beyond binary objectives to multi-class or free-form targets by replacing the F1-based task score, and incorporated into self-improving loops where learned verifier sets are coupled with RL-style methods such as GRPO as executable reward or filtering signals.

\bibliography{custom}

\begin{thebibliography}{32}
\providecommand{\natexlab}[1]{#1}

\bibitem[{Balunovi{\'c} et~al.(2025)Balunovi{\'c}, Dekoninck, Petrov, Jovanovi{\'c}, and Vechev}]{balunovic2025matharena}
Mislav Balunovi{\'c}, Jasper Dekoninck, Ivo Petrov, Nikola Jovanovi{\'c}, and Martin Vechev. 2025.
\newblock Matharena: Evaluating llms on uncontaminated math competitions.
\newblock \emph{arXiv preprint arXiv:2505.23281}.

\bibitem[{Comanici et~al.(2025)Comanici, Bieber, Schaekermann, Pasupat, Sachdeva, Dhillon, Blistein, Ram, Zhang, Rosen et~al.}]{comanici2025gemini}
Gheorghe Comanici, Eric Bieber, Mike Schaekermann, Ice Pasupat, Noveen Sachdeva, Inderjit Dhillon, Marcel Blistein, Ori Ram, Dan Zhang, Evan Rosen, and 1 others. 2025.
\newblock Gemini 2.5: Pushing the frontier with advanced reasoning, multimodality, long context, and next generation agentic capabilities.
\newblock \emph{arXiv preprint arXiv:2507.06261}.

\bibitem[{Ficek et~al.(2025)Ficek, Majumdar, Noroozi, and Ginsburg}]{ficek2025scoring}
Aleksander Ficek, Somshubra Majumdar, Vahid Noroozi, and Boris Ginsburg. 2025.
\newblock Scoring verifiers: Evaluating synthetic verification for code and reasoning.
\newblock \emph{arXiv preprint arXiv:2502.13820}.

\bibitem[{Gu et~al.(2024)Gu, Jiang, Shi, Tan, Zhai, Xu, Li, Shen, Ma, Liu et~al.}]{gu2024survey}
Jiawei Gu, Xuhui Jiang, Zhichao Shi, Hexiang Tan, Xuehao Zhai, Chengjin Xu, Wei Li, Yinghan Shen, Shengjie Ma, Honghao Liu, and 1 others. 2024.
\newblock A survey on llm-as-a-judge.
\newblock \emph{The Innovation}.

\bibitem[{Guo et~al.(2025)Guo, Yang, Zhang, Song, Wang, Zhu, Xu, Zhang, Ma, Bi et~al.}]{guo2025deepseek}
Daya Guo, Dejian Yang, Haowei Zhang, Junxiao Song, Peiyi Wang, Qihao Zhu, Runxin Xu, Ruoyu Zhang, Shirong Ma, Xiao Bi, and 1 others. 2025.
\newblock Deepseek-r1 incentivizes reasoning in llms through reinforcement learning.
\newblock \emph{Nature}, 645(8081):633--638.

\bibitem[{Huang et~al.()Huang, Chang, Lin, and Yang}]{huangbeyond}
Sian-Yao Huang, Li-Hsien Chang, Che-Yu Lin, and Cheng-Lin Yang.
\newblock Beyond oracle: Verifier-supervision for instruction hierarchy in reasoning and instruction-tuned llms.
\newblock In \emph{The Thirty-ninth Annual Conference on Neural Information Processing Systems}.

\bibitem[{Jain et~al.(2024)Jain, Han, Gu, Li, Yan, Zhang, Wang, Solar-Lezama, Sen, and Stoica}]{jain2024livecodebench}
Naman Jain, King Han, Alex Gu, Wen-Ding Li, Fanjia Yan, Tianjun Zhang, Sida Wang, Armando Solar-Lezama, Koushik Sen, and Ion Stoica. 2024.
\newblock Livecodebench: Holistic and contamination free evaluation of large language models for code.
\newblock \emph{arXiv preprint arXiv:2403.07974}.

\bibitem[{Kamoi et~al.(2024)Kamoi, Zhang, Zhang, Han, and Zhang}]{kamoi2024can}
Ryo Kamoi, Yusen Zhang, Nan Zhang, Jiawei Han, and Rui Zhang. 2024.
\newblock When can llms actually correct their own mistakes? a critical survey of self-correction of llms.
\newblock \emph{Transactions of the Association for Computational Linguistics}, 12:1417--1440.

\bibitem[{Khalifa et~al.(2025)Khalifa, Agarwal, Logeswaran, Kim, Peng, Lee, Lee, and Wang}]{khalifa2025process}
Muhammad Khalifa, Rishabh Agarwal, Lajanugen Logeswaran, Jaekyeom Kim, Hao Peng, Moontae Lee, Honglak Lee, and Lu~Wang. 2025.
\newblock Process reward models that think.
\newblock \emph{arXiv preprint arXiv:2504.16828}.

\bibitem[{Lee et~al.(2025{\natexlab{a}})Lee, Zeng, Jeong, Sohn, and Lee}]{lee2025correctly}
Chungpa Lee, Thomas Zeng, Jongwon Jeong, Jy-yong Sohn, and Kangwook Lee. 2025{\natexlab{a}}.
\newblock How to correctly report llm-as-a-judge evaluations.
\newblock \emph{arXiv preprint arXiv:2511.21140}.

\bibitem[{Lee et~al.(2025{\natexlab{b}})Lee, Kim, Kim, Cho, Kang, Kang, and Kim}]{lee2025checkeval}
Yukyung Lee, Joonghoon Kim, Jaehee Kim, Hyowon Cho, Jaewook Kang, Pilsung Kang, and Najoung Kim. 2025{\natexlab{b}}.
\newblock Checkeval: A reliable llm-as-a-judge framework for evaluating text generation using checklists.
\newblock In \emph{Proceedings of the 2025 Conference on Empirical Methods in Natural Language Processing}, pages 15782--15809.

\bibitem[{Li et~al.(2025)Li, Xu, Wang, Gong, Chen, Zhang, Wang, Lam, and Ji}]{li2025llms}
Songze Li, Chuokun Xu, Jiaying Wang, Xueluan Gong, Chen Chen, Jirui Zhang, Jun Wang, Kwok-Yan Lam, and Shouling Ji. 2025.
\newblock Llms cannot reliably judge (yet?): A comprehensive assessment on the robustness of llm-as-a-judge.
\newblock \emph{arXiv preprint arXiv:2506.09443}.

\bibitem[{Lightman et~al.(2023)Lightman, Kosaraju, Burda, Edwards, Baker, Lee, Leike, Schulman, Sutskever, and Cobbe}]{lightman2023let}
Hunter Lightman, Vineet Kosaraju, Yuri Burda, Harrison Edwards, Bowen Baker, Teddy Lee, Jan Leike, John Schulman, Ilya Sutskever, and Karl Cobbe. 2023.
\newblock Let's verify step by step.
\newblock In \emph{The twelfth international conference on learning representations}.

\bibitem[{Lou et~al.(2026)Lou, L{\'a}zaro-Gredilla, Dedieu, Wendelken, Lehrach, and Murphy}]{lou2026autoharness}
Xinghua Lou, Miguel L{\'a}zaro-Gredilla, Antoine Dedieu, Carter Wendelken, Wolfgang Lehrach, and Kevin~P Murphy. 2026.
\newblock Autoharness: improving llm agents by automatically synthesizing a code harness.
\newblock \emph{arXiv preprint arXiv:2603.03329}.

\bibitem[{Ma et~al.(2025)Ma, Liu, Luo, Huang, Zhu, and Che}]{ma2025advancing}
Zhiyuan Ma, Jiayu Liu, Xianzhen Luo, Zhenya Huang, Qingfu Zhu, and Wanxiang Che. 2025.
\newblock Advancing tool-augmented large language models via meta-verification and reflection learning.
\newblock In \emph{Proceedings of the 31st ACM SIGKDD Conference on Knowledge Discovery and Data Mining V. 2}, pages 2078--2089.

\bibitem[{Madaan et~al.(2023)Madaan, Tandon, Gupta, Hallinan, Gao, Wiegreffe, Alon, Dziri, Prabhumoye, Yang et~al.}]{madaan2023self}
Aman Madaan, Niket Tandon, Prakhar Gupta, Skyler Hallinan, Luyu Gao, Sarah Wiegreffe, Uri Alon, Nouha Dziri, Shrimai Prabhumoye, Yiming Yang, and 1 others. 2023.
\newblock Self-refine: Iterative refinement with self-feedback.
\newblock \emph{Advances in neural information processing systems}, 36:46534--46594.

\bibitem[{Pan et~al.(2026)Pan, Zou, Guo, Ni, and Zheng}]{pan2026natural}
Linyue Pan, Lexiao Zou, Shuo Guo, Jingchen Ni, and Hai-Tao Zheng. 2026.
\newblock Natural-language agent harnesses.
\newblock \emph{arXiv preprint arXiv:2603.25723}.

\bibitem[{Pei et~al.(2025)Pei, Du, and Jin}]{pei2025fover}
Yu~Pei, Yongping Du, and Xingnan Jin. 2025.
\newblock Fover: First-order logic verification for natural language reasoning.
\newblock \emph{Transactions of the Association for Computational Linguistics}, 13:1340--1359.

\bibitem[{Pronesti et~al.(2026)Pronesti, Belz, and Hou}]{pronesti2026beyond}
Massimiliano Pronesti, Anya Belz, and Yufang Hou. 2026.
\newblock Beyond outcome verification: Verifiable process reward models for structured reasoning.
\newblock \emph{arXiv preprint arXiv:2601.17223}.

\bibitem[{Pyatkin et~al.(2025)Pyatkin, Malik, Graf, Ivison, Huang, Dasigi, Lambert, and Hajishirzi}]{pyatkin2025generalizing}
Valentina Pyatkin, Saumya Malik, Victoria Graf, Hamish Ivison, Shengyi Huang, Pradeep Dasigi, Nathan Lambert, and Hannaneh Hajishirzi. 2025.
\newblock Generalizing verifiable instruction following.
\newblock \emph{arXiv preprint arXiv:2507.02833}.

\bibitem[{Singh et~al.(2025)Singh, Fry, Perelman, Tart, Ganesh, El-Kishky, McLaughlin, Low, Ostrow, Ananthram et~al.}]{singh2025openai}
Aaditya Singh, Adam Fry, Adam Perelman, Adam Tart, Adi Ganesh, Ahmed El-Kishky, Aidan McLaughlin, Aiden Low, AJ~Ostrow, Akhila Ananthram, and 1 others. 2025.
\newblock Openai gpt-5 system card.
\newblock \emph{arXiv preprint arXiv:2601.03267}.

\bibitem[{Snell et~al.(2024)Snell, Lee, Xu, and Kumar}]{snell2024scaling}
Charlie Snell, Jaehoon Lee, Kelvin Xu, and Aviral Kumar. 2024.
\newblock Scaling llm test-time compute optimally can be more effective than scaling model parameters.
\newblock \emph{arXiv preprint arXiv:2408.03314}.

\bibitem[{Wang et~al.(2025)Wang, Song, Zhu, Zhang, Yu, Chen, Song, Wang, Wang, Wu et~al.}]{wang2025trustjudge}
Yidong Wang, Yunze Song, Tingyuan Zhu, Xuanwang Zhang, Zhuohao Yu, Hao Chen, Chiyu Song, Qiufeng Wang, Cunxiang Wang, Zhen Wu, and 1 others. 2025.
\newblock Trustjudge: Inconsistencies of llm-as-a-judge and how to alleviate them.
\newblock \emph{arXiv preprint arXiv:2509.21117}.

\bibitem[{Wen et~al.(2024)Wen, Ke, Gu, Wu, Huang, Zhou, Li, Hu, Gao, Xu et~al.}]{wen2024benchmarking}
Bosi Wen, Pei Ke, Xiaotao Gu, Lindong Wu, Hao Huang, Jinfeng Zhou, Wenchuang Li, Binxin Hu, Wendy Gao, Jiaxin Xu, and 1 others. 2024.
\newblock Benchmarking complex instruction-following with multiple constraints composition.
\newblock \emph{Advances in Neural Information Processing Systems}, 37:137610--137645.

\bibitem[{Yang et~al.(2025)Yang, Jiang, He, Siu, Zhang, Liao, Li, Zeng, Jia, Wang et~al.}]{yang2025structeval}
Jialin Yang, Dongfu Jiang, Lipeng He, Sherman Siu, Yuxuan Zhang, Disen Liao, Zhuofeng Li, Huaye Zeng, Yiming Jia, Haozhe Wang, and 1 others. 2025.
\newblock Structeval: Benchmarking llms' capabilities to generate structural outputs.
\newblock \emph{arXiv preprint arXiv:2505.20139}.

\bibitem[{Yao et~al.(2022)Yao, Zhao, Yu, Du, Shafran, Narasimhan, and Cao}]{yao2022react}
Shunyu Yao, Jeffrey Zhao, Dian Yu, Nan Du, Izhak Shafran, Karthik~R Narasimhan, and Yuan Cao. 2022.
\newblock React: Synergizing reasoning and acting in language models.
\newblock In \emph{The eleventh international conference on learning representations}.

\bibitem[{Zhang et~al.(2026{\natexlab{a}})Zhang, Long, Bao, Feng, Zhang, Yue, and Wang}]{zhang2026memskill}
Haozhen Zhang, Quanyu Long, Jianzhu Bao, Tao Feng, Weizhi Zhang, Haodong Yue, and Wenya Wang. 2026{\natexlab{a}}.
\newblock Memskill: Learning and evolving memory skills for self-evolving agents.
\newblock \emph{arXiv preprint arXiv:2602.02474}.

\bibitem[{Zhang et~al.(2026{\natexlab{b}})Zhang, Fu, Xi, Jing, Chai, He, Zhang, Fan, An, Chen et~al.}]{zhang2026agentv}
Jiazheng Zhang, Ziche Fu, Zhiheng Xi, Wenqing Jing, Mingxu Chai, Wei He, Guoqiang Zhang, Chenghao Fan, Chenxin An, Wenxiang Chen, and 1 others. 2026{\natexlab{b}}.
\newblock Agentv-rl: Scaling reward modeling with agentic verifier.
\newblock \emph{arXiv preprint arXiv:2604.16004}.

\bibitem[{Zhang et~al.(2025{\natexlab{a}})Zhang, Hu, Upasani, Ma, Hong, Kamanuru, Rainton, Wu, Ji, Li et~al.}]{zhang2025agentic}
Qizheng Zhang, Changran Hu, Shubhangi Upasani, Boyuan Ma, Fenglu Hong, Vamsidhar Kamanuru, Jay Rainton, Chen Wu, Mengmeng Ji, Hanchen Li, and 1 others. 2025{\natexlab{a}}.
\newblock Agentic context engineering: Evolving contexts for self-improving language models.
\newblock \emph{arXiv preprint arXiv:2510.04618}.

\bibitem[{Zhang et~al.(2025{\natexlab{b}})Zhang, Zhu, Shen, Luo, Zhang, Liang, Yang, Lin, Qiao, Chen et~al.}]{zhang2025cfbench}
Tao Zhang, Chenglin Zhu, Yanjun Shen, Wenjing Luo, Yan Zhang, Hao Liang, Fan Yang, Mingan Lin, Yujing Qiao, Weipeng Chen, and 1 others. 2025{\natexlab{b}}.
\newblock Cfbench: A comprehensive constraints-following benchmark for llms.
\newblock In \emph{Proceedings of the 63rd Annual Meeting of the Association for Computational Linguistics (Volume 1: Long Papers)}, pages 32926--32944.

\bibitem[{Zhang et~al.(2026{\natexlab{c}})Zhang, Cui, Wang, Qiu, Li, Han, Huang, Qiu, Zhu, and He}]{zhang2026verified}
Xing Zhang, Yanwei Cui, Guanghui Wang, Qucy~Wei Qiu, Ziyuan Li, Fangwei Han, Yajing Huang, Hengzhi Qiu, Bin Zhu, and Peiyang He. 2026{\natexlab{c}}.
\newblock Verified multi-agent orchestration: A plan-execute-verify-replan framework for complex query resolution.
\newblock \emph{arXiv preprint arXiv:2603.11445}.

\bibitem[{Zhong et~al.(2025)Zhong, Du, Zhang, Hu, and Tang}]{zhong2025complexfuncbench}
Lucen Zhong, Zhengxiao Du, Xiaohan Zhang, Haiyi Hu, and Jie Tang. 2025.
\newblock Complexfuncbench: Exploring multi-step and constrained function calling under long-context scenario.
\newblock \emph{arXiv preprint arXiv:2501.10132}.

\end{thebibliography}

\appendix

\section{Prompts}
\label{sec:app-prompt}
We provide the overall verifier instruction, the prompts used for the modifier, critique, initial verifier-set generator, and context extractor, as well as the extra instruction added to the original prompt when verifiers are exposed as tools, in Prompts~\ref{prompt:vi}, \ref{prompt:mp}, \ref{prompt:cp}, \ref{prompt:gp}, \ref{prompt:cep}, and \ref{prompt:vt}, respectively.

\begin{prompt}[title={\footnotesize\texttt{VERIFIER-INSTRUCTION}}, label=prompt:vi]
You must output one or more Python modules in fenced ```python code blocks.\\
Each module must follow this exact contract:\\
\\
VERIFIER\_SPECS = [\\
    \{\\
        "name": "verifier\_name",\\
        "description": "what it checks",\\
        "requires": ["field\_a", "field\_b"]\\
    \},\\
    ...\\
]\\
\\
\# one Python function per verifier name\\
\# signature: verifier\_name(x, y, context=None)\\
\# each returns True or False\\
\# each verifier should use the context fields declared in VERIFIER\_SPECS when possible\\
\\
def verifier\_name(x, y, context=None):\\
    ...\\
    return True or False\\
\\
\# final accept/reject rule \\
\# signature: aggregate(checks, x, y, context=None)\\
\# checks is a dict: verifier\_name -> bool\\
\\
def aggregate(checks, x, y, context=None):\\
    return all(checks.values())\\
\\
Rules:\\
- deterministic Python only\\
- no network access\\
- no filesystem access\\
- no subprocesses\\
- allowed imports only from: math, re, json, statistics, fractions, decimal, itertools, ast, collections\\
- verifier sets should be small and interpretable\\
- do not hard-code labels from the dev set\\
- do not rely on exact gold-string matching\\
- do not parse raw y repeatedly inside every verifier if a context field can capture the needed information\\
- prefer generic checks reusable across tasks when possible, but task-specific checks are allowed when justified\\
- Do not use compile, exec, eval, open, input, globals, locals, vars, getattr, setattr, delattr, \_\_import\_\_, breakpoint, help, or any dynamic code execution/introspection. \\
\end{prompt}

\begin{prompt}[title={\footnotesize\texttt{Modifier Prompt}}, label=prompt:mp]
You are a verifier refiner. Produce improved child verifier bundles by editing the current bundle. \\
Each child should be a small deterministic Python verifier bundle following the required contract.\\
\\
Task description: \{task\_description\}\\
Current verifier bundle:\\```python\\\{node.program.source\_code\}\\```\\\\
Critic summary:\\\{critic\_summary\}\\\\
False positives:\\\{false\_positive\_examples\}\\\\
False negatives:\\\{false\_negative\_examples\}\\\\
Produce up to \{num\_children\} improved child verifier bundles.\\\\
\{VERIFIER-INSTRUCTION\}\\\\
Important refinement rules:\\
- Keep bundles small.\\
- Each verifier must declare a precise 'requires' list.\\
- Prefer improving or replacing weak checks rather than only adding more checks.\\
- Use context fields consistently.\\
- If a stronger context field is needed, add it to the requires list.\\
- Return only Python code blocks.
\end{prompt}

\begin{prompt}[title={\footnotesize\texttt{Critique Prompt}}, label=prompt:cp]
You are a verifier critic. Analyze where the current verifier set fails, \\
with special attention to false positives, false negatives, redundant checks, and missing context fields.\\
\\
Task description:\\\{task\_description\}\\\\
Current verifier code:\\```python\\\{node.program.source\_code\}\\```\\\\
Current metrics: PP=\{node.stats.pp\}, NP=\{node.stats.np\}, size=\{node.program.size\}\\\\
False positives:\\\{false\_positive\_examples\}\\\\
False negatives:\\\{false\_negative\_examples\}\\\\
Write a concise diagnosis with these sections:\\
1. Main loopholes causing false positives\\
2. Main over-restrictions causing false negatives\\
3. Redundant or weak checks\\
4. Missing verifier types or missing context fields\\
5. Suggested structured edits (ADD / REMOVE / REPLACE / MODIFY / CHANGE\_AGGREGATOR)\\
\end{prompt}

\begin{prompt}[title={\footnotesize\texttt{Initial Set Generator Prompt}}, label=prompt:gp]
You are an expert verifier-synthesis assistant.\\
Your job is to propose a small initial set of deterministic Python verifier bundles for judging whether a model output satisfies a target objective.\\
\\
Task description:\\\{task\_description\}\\\\
Need \{num\_seeds\} diverse initial verifier bundles.\\\\
Development examples (labeled):\\\{examples\}\\\\
\{VERIFIER-INSTRUCTION\}\\\\
Generate diverse, small verifier sets that are reusable across similar tasks.\\
Each verifier must declare a 'requires' list describing which context fields it needs.\\
Use context fields such as normalized\_text, final\_answer, steps, equations, json\_obj, citations, numbers, entities, or task-specific fields when justified.\\
Do not use compile, exec, eval, open, input, globals, locals, vars, getattr, setattr, delattr, \_\_import\_\_, breakpoint, help, or any dynamic code execution/introspection.\\
Do not include any extra explanation.\\
\end{prompt}

\begin{table*}[t!]
\centering
\small
\begin{tabularx}{\textwidth}{>{\raggedright\arraybackslash}p{0.24\textwidth}X}
\toprule
\textbf{Category} & \textbf{Description} \\
\midrule
Format / structure &
Checks output shape or structure, including formatting, sections, field layout, list structure, JSON structure, delimiters, required sections, and length/style wrappers. \\

Surface lexical &
Checks shallow text patterns such as keywords, regular expressions, exact phrases, token or string patterns, lexical overlap, and simple substring matches. \\

Entity / content presence &
Checks whether required entities, concepts, arguments, slots, labels, or other content items are present, without deeply validating their logical correctness. \\

Internal consistency &
Checks agreement or non-contradiction across parts of the same output, such as answer--explanation consistency, field-to-field consistency, cross-part compatibility, and self-consistency constraints. \\

Semantic / logical correctness proxy &
Checks deeper content validity using nontrivial heuristics or proxies, such as constraint satisfaction, logical coherence, semantic correctness proxies, and reasoning-pattern checks beyond simple lexical cues. \\

Lightweight symbolic / execution / numeric &
Checks correctness using executable, numeric, symbolic, or programmatic procedures, such as arithmetic checks, equation solving, code execution, parsing and evaluating expressions, or other formalized computations. \\

Anti-cheating / leakage control &
Checks for answer leakage, shortcut exploitation, template gaming, direct gold-string dependence, or other suspicious patterns intended to prevent verifier cheating. \\
\bottomrule
\end{tabularx}
\caption{\textbf{Verifier categories:} Each verifier function is assigned to exactly one category.}
\label{tab:verifier_taxonomy}
\end{table*}

\begin{prompt}[title={\footnotesize\texttt{Context Extractor Prompt}}, label=prompt:cep]
You build a JSON context for deterministic Python verifiers.\\
Given a task description, input x, model output y, and a list of required field names, return ONLY a JSON object.\\
\\
Rules:\\
- Return exactly one JSON object and nothing else.\\
- Include every required field exactly once.\\
- If a field cannot be extracted reliably, set it to null.\\
- Keep values concise but useful.\\
- Do not judge whether the output is correct overall.\\
- Extract structure from y; do not invent unsupported facts.\\
- Use evidence directly from x and y when possible.\\

Task description:\\\{task\_description\}\\
Input x:\\\{input\}\\\\
Model output y:\\\{output\}\\\\
Required fields:\\\{fields\}\\\\
Verifier specs:\\\{specs\}\\\\
Return ONLY a JSON object with exactly those required fields.\\
\end{prompt}

\begin{prompt}[title={\footnotesize\texttt{Verifiers as Tools System Prompt}}, label=prompt:vt]
...\\
Verifier tools may be available to help you check a draft solution before finalizing.\\
After you have a candidate solution, call at least one verifier tool by passing your current full draft solution.\\
Use verifier feedback to catch formatting, consistency, syntax, or contract issues when helpful. If a verifier flags a problem, revise your solution if appropriate.\\
Before giving the final answer, call at least one verifier tool on your current full draft solution.\\
\end{prompt}

\section{Additional Experimental Details}

After applying the benchmark-specific filtering described in Section~\ref{sec:exp-detail}, we retain 60 samples for AIME, 182 for LiveCodeBench, 240 for ComplexFuncBench, and 300 for IFBench. 
For \method, we tune the search hyperparameters using grid search over
\[
\alpha, \beta, \gamma \in \{0.1, 0.5, 1.0\}.
\]
Unless otherwise stated, all reported results use the best-performing hyperparameter configuration for each model-benchmark pair. Moreover, we use 10 randomly selected examples from each of the false-positive and false-negative sets.

\section{Verifier Taxonomy}

To analyze how the types of learned verifiers change after applying \method, we group verifier functions into the seven categories shown in Table~\ref{tab:verifier_taxonomy}. These categories are designed to cover both shallow surface-level checks and deeper semantic, logical, and executable verification behaviors.

\subsection{Example Initial and Final Verifier Sets}
For qualitative inspection, we provide the initial and final verifier sets for the best-performing model on each benchmark according to Table~\ref{tab:main_res}. In each case, the initial verifier set corresponds to the verifier bundle on the path from the root to the final selected node, while the final verifier set is the best verifier bundle returned by \method after search. The corresponding examples are shown for AIME in Figure~\ref{fig:AIME_verifier_sets}, LiveCodeBench in Figure~\ref{fig:LiveCodeBench_verifier_sets}, ComplexFuncBench in Figure~\ref{fig:ComplexFuncBench_verifier_sets}, and IFBench in Figure~\ref{fig:IFBench_verifier_sets}.
\begin{figure*}[p]
\centering
\adjustbox{valign=t,minipage=0.48\textwidth}{%
\begin{pyblock}{Root verifier set}
import re
from fractions import Fraction

VERIFIER_SPECS = [
{   
    'name': 'final_answer_present_and_parseable', 
    'description': 'Checks that a final answer can be
    extracted in a structured way from context or common
    answer markers.', 
    'requires': ['final_answer', 'normalized_text']
}, 
{   'name': 'internal_numeric_consistency', 
    'description': 'Checks that the extracted final 
    numeric answer agrees with at least one explicit
    numeric result stated near the end of the response.', 
    'requires': ['final_answer', 'normalized_text']
}, 
{
    'name': 'matches_reference_answer', 
    'description': 'Checks agreement with a provided
    reference answer when available.', 
    'requires': ['final_answer', 'normalized_text', 
     'reference_answer']
}]

def _text(context):
    return ((context or {}).get('normalized_text') or '')
    .strip()

def _extract_final(context):
    context = context or {}
    fa = context.get('final_answer')
    if fa is not None and str(fa).strip():
        return str(fa).strip()
    text = _text(context)
    boxed = re.findall('\\\\boxed\\{([^{}]+)\\}', text)
    if boxed:
        return boxed[-1].strip()
    m = re.search('(?:final answer|answer|therefore|thus|
    so)[:\\s]*([^\\n]+)', text, re.I)
    if m:
        return m.group(1).strip()
    return None

def _num_from_string(s):
    if s is None:
        return None
    s = str(s).replace(',', '').strip()
    m = re.search('-?\\d+(?:/\\d+)?(?:\\.\\d+)?', s)
    if not m:
        return None
    tok = m.group(0)
    try:
        if '/' in tok and '.' not in tok:
            return float(Fraction(tok))
        return float(tok)
    except Exception:
        return None

def final_answer_present_and_parseable(x, y, context=None):
    return _extract_final(context) is not None

def internal_numeric_consistency(x, y, context=None):
    text = _text(context)
    final_num = _num_from_string(_extract_final(context))
    if final_num is None: 
    return True
    lines = [ln.strip() for ln in text.splitlines() 
    if ln.strip()]
    tail = '\n'.join(lines[-8:]) if lines else text
    nums = [_num_from_string(m) for m in re.findall('-?\\
    d+(?:/\\d+)?(?:\\.\\d+)?', tail)]
    nums = [v for v in nums if v is not None]
    if not nums:
        return True
    return any((abs(v - final_num) <= 1e-06 for v in
    nums[:-1] if nums[:-1])) or len(nums) == 1

def _normalize(s):
    s = (s or '').strip().lower()
    s = re.sub('\\\\boxed\\{([^{}]+)\\}', '\\1', s)
    s = re.sub('[^a-z0-9/\\.\\-]+', '', s)
    return s

def matches_reference_answer(x, y, context=None):
    context = context or {}
    ref = context.get('reference_answer')
    if ref is None or str(ref).strip() == '':
        return False
\end{pyblock}
}%
\hfill
\adjustbox{valign=t,minipage=0.48\textwidth}{%
\begin{pyblock}{Best verifier set}
import re
from fractions import Fraction

VERIFIER_SPECS = [
{
    "name": "reference_supported_final_answer",
    "description": "Requires an extractable final 
    answer that semantically matches the reference answer.",
    "requires": ["final_answer", "normalized_text", 
    "reference_answer"]
},
{
    "name": "no_multiple_distinct_final_claims",
    "description": "Rejects outputs with multiple
    distinct final claims or boxed answers.",
    "requires": ["final_answer", "normalized_text"]
},
{
    "name": "no_obvious_unfinished_or_repetitive_reasoning",
    "description": "Rejects outputs with obvious
    unresolved search language or extreme repetition.",
    "requires": ["normalized_text"]
}]

def _text(context):
    return ((context or {}).get("normalized_text") or "")
    .strip()

def _normalize(s):
    s = s or ""
    s = re.sub(r'\\boxed\{([^{}]+)\}', r'\1', s)
    s = re.sub(r'\\left|\\right', '', s)
    s = s.lower().strip()
    s = re.sub(r'\s+', '', s)
    s = re.sub(r'[^a-z0-9/\.\-\\{}]', '', s)
    return s

def _extract_final(context):
    context = context or {}
    fa = context.get("final_answer")
    if fa is not None and str(fa).strip():
        return str(fa).strip()
    text = _text(context)
    boxed = re.findall(r'\\boxed\{([^{}]+)\}', text)
    if boxed:
        return boxed[-1].strip()
    matches = re.findall(r'(?i)(?:final answer|answer|
    therefore|thus|hence)\s*(?:is|=|:)?\s*([^\n]{1,80})',
    text)
    return matches[-1].strip() if matches else None

def _num(s):
    if s is None:
        return None
    s = str(s).strip().replace(",", "")
    m = re.fullmatch(r'\\frac\{(-?\d+)\}\{(-?\d+)\}', s)
    if m and int(m.group(2)) != 0:
        return float(Fraction(int(m.
        group(1)), 
        int(m.group(2))))
    m = re.fullmatch(r'(-?\d+)\s*/\s*(-?\d+)', s)
    if m and int(m.group(2)) != 0:
        return float(Fraction(int(m.group(1)), 
        int(m.group(2))))
    m = re.search(r'-?\d+(?:/\d+)?(?:\.\d+)?', s)
    if not m:
        return None
    tok = m.group(0)
    try:
        if "/" in tok and "." not in tok:
            a, b = tok.split("/")
            if int(b) == 0:
                return None
            return float(Fraction(int(a), int(b)))
        return float(tok)
    except Exception:
        return None

def _candidate_claims(text):
    cands = []
    cands.extend([s.strip() for s in re.findall(r'\\boxed\
    {([^{}]+)\}', text)])
    cands.extend([s.strip() for s in re.findall(r'(?i)
    (?:final answer|answer)\s*(?:is|=|:)?\s*([^\n]{1,80})',
    text)])
    return [c for c in cands if c]

def reference_supported_final_answer(x, y, context=None):
    context = context or {}
\end{pyblock}
}%
\end{figure*}

\begin{figure*}[t]
\centering
\adjustbox{valign=t,minipage=0.48\textwidth}{%
\begin{pyblock}{Root verifier set (continued)}
    cand = _extract_final(context)
    if cand is None:
        return False
    a = _num_from_string(cand)
    b = _num_from_string(ref)
    if a is not None and b is not None:
        tol = context.get('numeric_tolerance', 1e-06)
        return abs(a - b) <= tol
    return _normalize(cand) == _normalize(str(ref))

def aggregate(checks, x, y, context=None):
    return checks.get('final_answer_present_and_parseable',
    False) and checks.get('matches_reference_answer', 
    False) and checks.get('internal_numeric_consistency', 
    False)

\end{pyblock}
}%
\hfill
\adjustbox{valign=t,minipage=0.48\textwidth}{%
\begin{pyblock}{Best verifier set (continued)}
    ref = context.get("reference_answer")
    ans = _extract_final(context)
    if ref is None or str(ref).strip() == "" or ans is 
    None:
        return False
    a = _num(ans)
    b = _num(str(ref))
    if a is not None and b is not None:
        return abs(a - b) <= context.get("numeric_
        tolerance", 1e-6)
    return _normalize(ans) == _normalize(str(ref))

def no_multiple_distinct_final_claims(x, y, context=None):
    text = _text(context)
    cands = _candidate_claims(text)
    if not cands:
        return False
    vals = []
    for c in cands[-6:]:
        n = _num(c)
        vals.append(("n", round(n, 10)) if n is not None 
        else ("t", _normalize(c)))
    return len(set(vals)) <= 1

def no_obvious_unfinished_or_repetitive_reasoning
(x, y, context=None):
    text = _text(context).lower()
    if not text:
        return False
    if text.count("already tried") >= 5:
        return False
    if len(re.findall(r'\btry\b', text)) >= 12:
        return False
    if "let's try" in text and "\\boxed" not in text and
    "final answer" not in text and "answer" not in text:
        return False
    return True

def aggregate(checks, x, y, context=None):
    return all(checks.values())
\end{pyblock}
}%
\caption{Example verifier sets for AIME. Left: the initial verifier set. Right: the best verifier set selected by \method.}
\label{fig:AIME_verifier_sets}
\end{figure*}

\begin{figure*}[p]
\centering
\adjustbox{valign=t,minipage=0.48\textwidth}{%
\begin{pyblock}{Root verifier set}
VERIFIER_SPECS = [
{
    'name': 'code_response_present', 
    'description': 'Checks that the response contains 
    substantial Python-like code rather than prose.',
    'requires': ['normalized_text']
}, 
{
    'name': 'structural_task_match', 
    'description': 'Checks that the response matches the 
    expected task structure, including class Solution
    /method for method tasks and plausible solve/main flow
    for stdin tasks.', 
    'requires': ['normalized_text', 'task_type', 
    'task_method_name', 'expects_class_solution', 
    'starter_code']
}, 
{
    'name': 'no_obvious_malformed_or_truncated_code', 
    'description': 'Rejects responses with obvious 
    malformed/truncated structure or extreme repeated
    boilerplate.', 
    'requires': ['normalized_text']
}, 
{
    'name': 'constraint_aware_bruteforce_filter', 
    'description': 'Uses generic constraint-aware
    heuristics to reject likely brute-force or 
    cubic/quadratic approaches on large instances.', 
    'requires': ['normalized_text', 'constraints_text', 
    'full_problem_text']
}]

import re

def _txt(context, y):
    text = (context or {}).get('normalized_text', y or '')
    return text if isinstance(text, str) else ''

def _clean(text):
    t = text.strip()
    if t.startswith('```'):
        lines = t.splitlines()
        if lines and lines[0].startswith('```'):
            lines = lines[1:]
        if lines and lines[-1].strip() == '```':
            lines = lines[:-1]
        t = '\n'.join(lines)
    return t.strip()

def _large_problem(context):
    blob = ' '.join([str((context or {}).get('constraints_
    text', '') or ''), str((context or {}).get('full_
    problem_text', '') or '')]).replace(',', '').lower()
    if re.search('10\\s*\\^\\s*[5-9]\\b', blob):
        return True
    if re.search('10\\s*\\^\\s*1[0-9]\\b', blob):
        return True
    nums = [int(x) for x in re.findall('\\b\\d+\\b', blob)]
    return any((v >= 100000 for v in nums))

def code_response_present(x, y, context=None):
    text = _clean(_txt(context, y))
    if not text:
        return False
    low = text.lower()
    if any((p in low for p in ['your code here', 
    'not implemented', 'placeholder'])):
        return False
    lines = [ln.strip() for ln in text.splitlines() 
    if ln.strip()]
    if len(lines) < 2:
        return False
    codeish = sum((1 for ln in lines if re.search(
    '\\b(def|class|import|for|while|if|return|print)\\b
    |[=:()\\[\\]]', ln)))
    return codeish >= 2

def structural_task_match(x, y, context=None):
    context = context or {}
    text = _clean(_txt(context, y))
    low = text.lower()
    task_type = context.get('task_type')
    method = context.get('task_method_name')
    expects_class = bool(context.get('expects_class_
    solution'))
    starter = str(context.get('starter_code', '') or '')
    .lower()
    method_task = task_type == 'method' or method or 
    expects_class or ('class solution' in starter)
    if method_task:
        if expects_class or 'class solution' in starter:
            if not re.search('^\\s*class\\s+solution\\s*
            [:(]', text, re.I | re.M):
\end{pyblock}
}%
\hfill
\adjustbox{valign=t,minipage=0.48\textwidth}{%
\begin{pyblock}{Best verifier set}
VERIFIER_SPECS = [
{
    "name": "code_response_present",
    "description": "Checks that the response contains s
    ubstantial Python-like code rather than prose.",
    "requires": ["normalized_text"]
},
{
    "name": "python_parseable_basic",
    "description": "Uses parse metadata when available, 
    otherwise conservative syntax sanity checks.",
    "requires": ["normalized_text", "python_ast_parse_ok"]
},
{
    "name": "task_contract_compatible",
    "description": "Checks that the answer matches the 
    expected method/class or stdin/stdout contract.",
    "requires": ["normalized_text", "task_type",
    "task_method_name", "expects_class_solution", 
    "starter_code"]
},
{
    "name": "small_oracle_or_sample_signal",
    "description": "Uses any available execution-based 
    correctness signal such as samples or tiny-oracle 
    checks.",
    "requires": ["sample_tests_passed", "sample_tests
    _total", "tiny_oracle_passed", "tiny_oracle_total",
    "can_execute"]
},
{
    "name": "no_suspicious_uncertainty_comments",
    "description": "Rejects responses with strong
    uncertainty/admission comments that often correlate 
    with incorrect code.",
    "requires": ["normalized_text"]
}]

import re

def _txt(context, y):
    text = (context or {}).get("normalized_text", y or "")
    return text if isinstance(text, str) else ""

def _clean(text):
    t = text.strip()
    if t.startswith("```"):
        lines = t.splitlines()
        if lines and lines[0].startswith("```"):
            lines = lines[1:]
        if lines and lines[-1].strip() == "```":
            lines = lines[:-1]
        t = "\n".join(lines)
    return t.strip()

def code_response_present(x, y, context=None):
    text = _clean(_txt(context, y))
    if not text:
        return False
    low = text.lower()
    if any(p in low for p in ["your code here", "not 
    implemented", "placeholder"]):
        return False
    lines = [ln.strip() for ln in text.splitlines() if 
    ln.strip()]
    if len(lines) < 2:
        return False
    codeish = sum(
        1
        for ln in lines
        if re.search(r"\b(def|class|import|from|for|while|
        if|return|print)\b|[=:()\[\]]", ln)
    )
    return codeish >= 2

def python_parseable_basic(x, y, context=None):
    context = context or {}
    parse_ok = context.get("python_ast_parse_ok")
    if isinstance(parse_ok, bool):
        return parse_ok

    text = _clean(_txt(context, y))
    if not text:
        return False
    lines = [ln.rstrip() for ln in text.splitlines() if 
    ln.strip()]
    if not lines:
        return False
    tail = lines[-1].strip()
    if re.search(r"\b(def|class|if|elif|else|for|while
    |try|except|with)\b[^:]*\$", tail):
        return False
    if re.search(r"[=+\-*/,(]\s*\$", tail):
        return False
\end{pyblock}
}%
\end{figure*}

\begin{figure*}[t]
\centering
\adjustbox{valign=t,minipage=0.48\textwidth}{%
\begin{pyblock}{Root verifier set (continued)}
                return False
        if method and (not re.search('\\bdef\\s+' + 
        re.escape(str(method)) + '\\s*\\(', text)):
            return False
        return True
    if task_type == 'stdin_stdout':
        has_input = bool(re.search('\\binput\\s*\\(|sys
        \\.stdin|stdin\\.read|stdin\\.readline|open\\s*
        \\(\\s*0\\s*\\)', low))
        has_output = bool(re.search('\\bprint\\s*\\(|sys
        \\.stdout\\.write', low))
        return has_input and has_output
    return True

def no_obvious_malformed_or_truncated_code(x, y, 
context=None):
    text = _clean(_txt(context, y))
    if not text:
        return False
    lines = [ln.rstrip() for ln in text.splitlines() if 
    ln.strip()]
    if not lines:
        return False
    repeated = {}
    for ln in lines:
        s = ln.strip()
        if len(s) >= 10:
            repeated[s] = repeated.get(s, 0) + 1
    if repeated and max(repeated.values()) >= 10:
        return False
    tail = lines[-1].strip().lower()
    if re.search('\\b(if|for|while|def|class|elif|else|
    try|except|with)\\b[^:]*\$', tail):
        return False
    if re.search('[=+\\-*/,(]\\s*\$', tail):
        return False
    opens = text.count('(') + text.count('[') + text.
    count('{')
    closes = text.count(')') + text.count(']') + text.
    count('}')
    if opens != closes:
        return False
    return True

def constraint_aware_bruteforce_filter(x, y, context=None):
    context = context or {}
    text = _clean(_txt(context, y))
    low = text.lower()
    if not _large_problem(context):
        return True
    if low.count('for ') >= 3:
        return False
    if re.search('for\\s+\\w+\\s+in\\s+range\\s*\\(\\s*n\\
    s*\\).*for\\s+\\w+\\s+in\\s+range\\s*\\(', low, re.S):
        return False
    if re.search('for\\s+\\w+\\s+in\\s+range\\s*\\(\\s*m\\
    s*\\).*for\\s+\\w+\\s+in\\s+range\\s*\\(', low, re.S):
        return False
    if re.search('for\\s+\\w+\\s+in\\s+range\\s*\\(\\s*q\\
    s*\\).*for\\s+\\w+\\s+in\\s+range\\s*\\(', low, re.S):
        return False
    if re.search('for\\s+\\w+\\s+in\\s+range\\s*\\(\\s*-?
    max_|for\\s+\\w+\\s+in\\s+range\\s*\\(\\s*-?r', low):
        return False
    if re.search('\\[[^\\]]+:[^\\]]+\\]', low) and 'for ' 
    in low:
        return False
    if 'for last_diff in dp[j]' in low:
        return False
    return True

def aggregate(checks, x, y, context=None):
    return all(checks.values())

\end{pyblock}
}%
\hfill
\adjustbox{valign=t,minipage=0.48\textwidth}{%
\begin{pyblock}{Best verifier set (continued)}
    return True

def task_contract_compatible(x, y, context=None):
    context = context or {}
    text = _clean(_txt(context, y))
    low = text.lower()
    task_type = context.get("task_type")
    method = context.get("task_method_name")
    expects_class = bool(context.get("expects_
    class_solution"))
    starter = str(context.get("starter_code", "") or "")
    .lower()

    method_task = task_type == "method" or bool(method) or 
    expects_class or "class solution" in starter
    if method_task:
        if expects_class or "class solution" in starter:
            if not re.search(r"^\s*class\s+solution\s*[:(]
            ", text, re.I | re.M):
                return False
        if method and not re.search(r"\bdef\s+" + re.
        escape(str(method)) + r"\s*\\(\\", text):
            return False
        return True

    if task_type == "stdin_stdout":
        return bool(
            re.search(r"\binput\s*\\(|sys\.stdin|stdin\
            .read|stdin\.readline|open\s*\(\s*0\s*\)", low)
            or re.search(r"\bprint\s*\\(|sys\.stdout\
            .write", low)
            or re.search(r"\bdef\s+(solve|main)\s*\\(",
            low)
            or "__name__" in low)
    return True

def small_oracle_or_sample_signal(x, y, context=None):
    context = context or {}
    can_execute = context.get("can_execute")

    tp = context.get("tiny_oracle_passed")
    tt = context.get("tiny_oracle_total")
    if can_execute and isinstance(tt, int) and tt > 0 and 
    isinstance(tp, int):
        return tp == tt

    sp = context.get("sample_tests_passed")
    st = context.get("sample_tests_total")
    if can_execute and isinstance(st, int) and st > 0 and 
    isinstance(sp, int):
        return sp == st

    return True

def no_suspicious_uncertainty_comments(x, y, context=None):
    text = _clean(_txt(context, y)).lower()
    if not text:
        return False

    bad_patterns = [
        r"\btoo slow\b",
        r"\bnot sure\b",
        r"\bunsure\b",
        r"\bprobably\b",
        r"\bmaybe\b",
        r"\blet'?s try\b",
        r"\bthat'?s complicated\b",
        r"\bbut that'?s too slow\b",
    ]
    hits = sum(1 for p in bad_patterns if re.search(p, 
    text))
    return hits == 0

def aggregate(checks, x, y, context=None):
    required = [
        "code_response_present",
        "python_parseable_basic",
        "task_contract_compatible",
        "small_oracle_or_sample_signal",
    ]
    for k in required:
        if not checks.get(k, False):
            return False
    return checks.get("no_suspicious_uncertainty_comments"
    , True)
\end{pyblock}
}%
\caption{Example verifier sets for LiveCodeBench. Left: the initial verifier set. Right: the best verifier set selected by \method.}
\label{fig:LiveCodeBench_verifier_sets}
\end{figure*}

\begin{figure*}[p]
\centering
\adjustbox{valign=t,minipage=0.48\textwidth}{%
\begin{pyblock}{Root verifier set}
VERIFIER_SPECS = [
{
    'name': 'ends_with_assistant_text', 
    'description': 'Checks if the final message is from 
    the assistant and contains text rather than a function
    call.', 
    'requires': ['messages']
}, 
{
    'name': 'check_expected_functions_called', 
    'description': 'Checks if required tools corresponding
    to specific user requests are called successfully at 
    least once.', 
    'requires': ['messages']
}]

def ends_with_assistant_text(x, y, context=None):
    import json
    messages = context.get('messages') if context and
    'messages' in context else None
    if messages is None:
        try:
            messages = json.loads(y) if isinstance(y, str)
            else y
        except Exception:
            messages = []
    if not messages:
        return False
    last_msg = messages[-1]
    if last_msg.get('role') != 'assistant':
        return False
    content = last_msg.get('content')
    if not content or not isinstance(content, str):
        return False
    if last_msg.get('function_call'):
        return False
    return True

def check_expected_functions_called(x, y, context=None):
    import json
    import re
    messages = context.get('messages') if context and 
    'messages' in context else None
    if messages is None:
        try:
            messages = json.loads(y) if isinstance(y, str) 
            else y
        except Exception:
            messages = []
    available_funcs = set(re.findall('Function `([a-zA-Z0-
    9_]+)`', x))
    user_request_match = re.search('User request:\\n(.*?)
    \\n\\nAvailable functions', x, re.DOTALL)
    if user_request_match:
        user_request = user_request_match.group(1).lower()
    else:
        user_request = x.lower()
    expectations = [(['details of the fastest', 'details 
    of the cheapest', 'flight details', 'ticket details'],
    ['Get_Flight_Details']), (['activit', 'attraction', 
    'place to visit', 'places to visit'], ['Search_
    Attractions']), (['hotel', 'accommodation', 'stay'],
    ['Search_Hotels', 'Search_Hotels_By_Coordinates']),
    (['flight', 'plane ticket', 'airline'], ['Search_
    Flights']), (['taxi', 'cab', 'ride'], ['Search_Taxi'])
    , (['car rental', 'rent a car', 'rental car'],
    ['Search_Car_Rentals']), (['available rooms', 'room 
    availability'], ['Get_Room_List_With_Availability'])]
    successful_calls = set()
    for (i, msg) in enumerate(messages):
        if msg.get('role') == 'assistant' and 
        'function_call' in msg:
            calls = msg.get('function_call', [])
            if isinstance(calls, dict):
                calls = [calls]
            if i + 1 < len(messages) and messages[i + 1]
            .get('role') == 'observation':
                obs_msg = messages[i + 1]
                obs_content = obs_msg.get('content', [])
                if isinstance(obs_content, str):
                    try:
                        obs_content = json.loads
                        (obs_content)
                    except Exception:
                        pass
                if isinstance(obs_content, list):
                    for (call, obs) in zip(calls, 
                    obs_content):
                        func_name = call.get('name')
                        obs_text = ''
                        if isinstance(obs, dict):
                            obs_text = str(obs.get
                            ('content', '')) + str(obs)

\end{pyblock}
}%
\hfill
\adjustbox{valign=t,minipage=0.48\textwidth}{%
\begin{pyblock}{Best verifier set}
VERIFIER_SPECS = [
{
    "name": "ends_with_assistant_text",
    "description": "Checks if the final message is from 
    the assistant and contains text rather than a function 
    call.",
    "requires": ["messages"]
},
{
    "name": "check_expected_functions_called",
    "description": "Checks if required tools corresponding
    to specific user requests are called successfully at
    least once.",
    "requires": ["messages"]
},
{
    "name": "check_no_clarification_or_refusal",
    "description": "Rejects outputs where the assistant
    gives up, asks for basic missing parameters, or blames
    technical issues.",
    "requires": ["messages"]
},
{
    "name": "check_no_hallucinated_success",
    "description": "Rejects outputs that end with a final 
    text response immediately after an API error, ensuring 
    the model retries instead of hallucinating success or
    giving up.",
    "requires": ["messages"]
}]

import json
import re

def ends_with_assistant_text(x, y, context=None):
    messages = context.get("messages") if context and
    "messages" in context else None
    if messages is None:
        try:
            messages = json.loads(y) if isinstance(y, str)
            else y
        except Exception:
            messages = []            
    if not messages:
        return False        
    last_msg = messages[-1]
    if last_msg.get("role") != "assistant":
        return False        
    content = last_msg.get("content")
    if not content or not isinstance(content, str):
        return False        
    if last_msg.get("function_call"):
        return False        
    return True

def check_expected_functions_called(x, y, context=None):
    messages = context.get("messages") if context and 
    "messages" in context else None
    if messages is None:
        try:
            messages = json.loads(y) if isinstance(y, str) 
            else y
        except Exception:
            messages = []            
    available_funcs = set(re.findall(r'Function `([a-zA-Z0
    -9_]+)`', x))    
    user_request_match = re.search(r'User request:\n(.*?)
    \n\nAvailable functions', x, re.DOTALL)
    if user_request_match:
        user_request = user_request_match.group(1).lower()
    else:
        user_request = x.lower()        
    expectations = [
        (["details of the fastest", "details of the 
        cheapest", "flight details", "ticket details"], 
        ["Get_Flight_Details"]), (["activit", "attraction",
        "place to visit", "places to visit"],["Search_
        Attractions"]), (["book a hotel", "hotel 
        availability", "reserve a hotel", "find a hotel", 
        "stay at the ", "stay in a hotel", "stay at "],
        ["Search_Hotels", "Search_Hotels_By_Coordinates"])
        , (["flight", "plane ticket", "airline", "air 
        ticket"], ["Search_Flights", "Search_Flights_Multi
        _Stops"]), (["taxi", "cab", "ride"], ["Search_Taxi
        "]), (["car rental", "rent a car", "rental car"],
        ["Search_Car_Rentals"]), (["available rooms", 
        "room availability"], ["Get_Room_List_With_
        Availability"])]
    successful_calls = set()
    for i, msg in enumerate(messages):
        if msg.get("role") == "assistant" and "function_
        call" in msg:
            calls = msg.get("function_call", [])

\end{pyblock}
}%
\end{figure*}

\begin{figure*}[t]
\centering
\adjustbox{valign=t,minipage=0.48\textwidth}{%
\begin{pyblock}{Root verifier set (continued)}
                        elif isinstance(obs, str):
                            obs_text = obs
                        if 'problem with your api call' 
                        not in obs_text.lower():
                            successful_calls.add(func_name)
                elif isinstance(obs_content, dict):
                    if calls:
                        func_name = calls[0].get('name')
                        obs_text = str(obs_content.get
                        ('content', '')) + str(obs_content)
                        if 'problem with your api call'
                        not in obs_text.lower():
                            successful_calls.add(func_name)
                elif isinstance(obs_content, str):
                    if calls:
                        func_name = calls[0].get('name')
                        if 'problem with your api call'
                        not in obs_content.lower():
                            successful_calls.add(func_name)
    for (keywords, funcs) in expectations:
        if any((kw in user_request for kw in keywords)):
            applicable_funcs = [f for f in funcs if f in 
            available_funcs]
            if not applicable_funcs:
                continue
            if not any((f in successful_calls for f in
            applicable_funcs)):
                return False
    return True

def aggregate(checks, x, y, context=None):
    return all(checks.values())

\end{pyblock}
}%
\hfill
\adjustbox{valign=t,minipage=0.48\textwidth}{%
\begin{pyblock}{Best verifier set (continued)}
            if isinstance(calls, dict):
                calls = [calls]            
            if i + 1 < len(messages) and messages[i+1].get
            ("role") == "observation":
                obs_msg = messages[i+1]
                obs_content = obs_msg.get("content", [])
                if isinstance(obs_content, str):
                    try:
                        obs_content = json.loads(obs_
                        content)
                    except Exception: pass                
                if isinstance(obs_content, list):
                    for call, obs in zip(calls, obs_
                    content):
                        func_name = call.get("name")
                        obs_text = str(obs).lower()
                        if "problem with your api call" 
                        not in obs_text:
                            successful_calls.add(func_name)
                elif isinstance(obs_content, dict):
                    if calls:
                        func_name = calls[0].get("name")
                        obs_text = str(obs_content).lower()
                        if "problem with your api call" 
                        not in obs_text:
                            successful_calls.add(func_name)
                elif isinstance(obs_content, str):
                    if calls:
                        func_name = calls[0].get("name")
                        if "problem with your api call" 
                        not in obs_content.lower():
                            successful_calls.add(func_name)
    for keywords, funcs in expectations:
        if any(kw in user_request for kw in keywords):
            applicable_funcs = [f for f in funcs if f in 
            available_funcs]
            if not applicable_funcs: continue            
            if not any(f in successful_calls for f in 
            applicable_funcs):
                return False          
    return True

def check_no_clarification_or_refusal(x, y, context=None):
    messages = context.get("messages") if context and 
    "messages" in context else None
    if messages is None:
        try:
            messages = json.loads(y)
        except Exception:
            messages = []                   
    last_msg = messages[-1]     
    content = last_msg.get("content")       
    content_lower = content.lower()
    refusal_phrases = ["need a bit more information",
    "could you provide", "please clarify",
    "technical issue", "i need more information",
    "could you please clarify", "recommend booking 
    directly", "try again shortly", "unable to book",
    "unable to find", "could not find", "couldn't find", 
    "once you provide"]
    if any(phrase in content_lower for phrase in refusal_
    phrases): return False        
    return True

def check_no_hallucinated_success(x, y, context=None):
    messages = context.get("messages") if context and 
    "messages" in context else None
    if messages is None:
        try:
            messages = json.loads(y)
        except Exception:
            messages = []            
    for i, msg in enumerate(messages):
        if msg.get("role") == "observation":
            content = msg.get("content", [])
            obs_text = str(content).lower()
            if "problem with your api call" in obs_text:
                has_subsequent_call = False
                for subsequent_msg in messages[i+1:]:
                    if subsequent_msg.get("role") == 
                    "assistant" and "function_call" in 
                    subsequent_msg:
                        has_subsequent_call = True
                        break
                if not has_subsequent_call: return False
    return True

def aggregate(checks, x, y, context=None):
    return all(checks.values())

\end{pyblock}
}%
\caption{Example verifier sets for ComplexFuncBench. Left: the initial verifier set. Right: the best verifier set selected by \method.}
\label{fig:ComplexFuncBench_verifier_sets}
\end{figure*}

\begin{figure*}[p]
\centering
\adjustbox{valign=t,minipage=0.48\textwidth}{%
\begin{pyblock}{Root verifier set}
import re
VERIFIER_SPECS = [
{
    'name': 'verify_instruction_constraints', 
    'description': 'Consolidated verifier for IFEval-style 
    constraints, expanded to handle word bounds, list 
    sampling, conjunctions, punctuation, custom bullets, 
    n-th word rules, prose format, unique words, exact 
    echoing, repetition mutations, emojis, stairs 
    indentation, sentence counts, char limits, consecutive 
    first letters, word per line, intro-then-bullet
    structures, options, no explanation, nested
    brackets/quotes, and sub-bullets.', 
    'requires': []
}]

def get_words(text):
    words = text.split()
    PUNCT = '.,!?"\'()[]{}<>:;*-\_~`#|\\/="'--'
    cleaned = [w.strip(PUNCT) for w in words]
    return [w for w in cleaned if w]

def verify_instruction_constraints(x, y, context=None):
    query = x if isinstance(x, str) else x.get('Query', '')
    output = y if isinstance(y, str) else y.get('Output',
    '')
    pattern_freq = 'keyword\\s+[\'\\"]?([A-Za-z0-9_-]+)
    [\'\\"]?\\s+([a-z0-9\\s]+?)\\s+in your response'
    for match in re.finditer(pattern_freq, query, flags=re
    .IGNORECASE):
        word = match.group(1).lower()
        count_phrase = match.group(2).lower().strip()
        expected_count = None
        number_map = {'once': 1, 'twice': 2, 'three times'
        :3, 'four times': 4, 'five times': 5, 'six times':
        6, 'seven times': 7, 'eight times': 8, 
        'nine times': 9, 'ten times': 10}
        if count_phrase in number_map:
            expected_count = number_map[count_phrase]
        else:
            num_match = re.search('\\d+', count_phrase)
            if num_match:
                expected_count = int(num_match.group())
        if expected_count is not None:
            actual_count = len(re.findall('\\b' + re
            .escape(word) + '\\b', output, flags=re
            .IGNORECASE))
            if actual_count != expected_count:
                return False
    pattern_sent = 'include keyword\\s+[\'\\"]?([A-Za-z0-9
    _-]+)[\'\\"]?\\s+in the\\s+(\\d+)-(?:st|nd|rd|th)\\s+
    sentence(?:,\\s+as the\\s+(\\d+)
    -(?:st|nd|rd|th)\\s+word)?'
    for match in re.finditer(pattern_sent, query, re
    .IGNORECASE):
        word = match.group(1).lower()
        sent_idx = int(match.group(2)) - 1
        word_idx = int(match.group(3)) - 1 if match.group
        (3) else None
        sentences = [s.strip() for s in re.split('(?<=[.!
        ?])\\s+|\\n+', output.strip()) if s.strip()]
        if sent_idx >= len(sentences):
            return False
        target_sent = sentences[sent_idx]
        words_list = get_words(target_sent)
        if word_idx is not None:
            if word_idx >= len(words_list):
                return False
            if words_list[word_idx].lower() != word:
                return False
        elif word not in [w.lower() for w in words_list]:
            return False
    pattern_global = 'The second word in your response and 
    the second to last word in your response should be the
    word\\s+[\'\\"]?([A-Za-z0-9_-]+)[\'\\"]?'
    for match in re.finditer(pattern_global, query,
    re.IGNORECASE):
        target_word = match.group(1).lower()
        words_list = get_words(output)
        if len(words_list) < 2:
            return False
        if words_list[1].lower() != target_word or
        words_list[-2].lower() != target_word:
            return False
    pattern_num = 'Include exactly\\s+(\\d+)\\s+
    numbers? in the response'
    for match in re.finditer(pattern_num, query, re
    .IGNORECASE):
        expected_count = int(match.group(1))
        actual_count = len(re.findall('\\b\\d+\\b',
        output))
        if actual_count != expected_count:
            return False
\end{pyblock}
}%
\hfill
\adjustbox{valign=t,minipage=0.48\textwidth}{%
\begin{pyblock}{Best verifier set}
import re

VERIFIER_SPECS = [
{
    "name": "verify_instruction_constraints",
    "description": "Consolidated verifier for IFEval-
    style constraints, expanded to handle word bounds,
    list sampling, conjunctions, punctuation, custom 
    bullets, n-th word rules, prose format, unique 
    words, exact echoing, repetition mutations, emojis,
    stairs indentation, sentence counts, char limits, 
    consecutive first letters, word per line, intro-
    then-bullet structures, options, no explanation, 
    nested brackets/quotes, sub-bullets, trigram 
    overlaps, and sentence type ratios.",
    "requires": []
}]

def get_words(text):
    # Remove emojis before splitting to prevent them from
    # counting as words or shifting position bounds
    emoji_pattern = r'[\u2300-\u2BFF\U00010000-\U0010FFFF]'
    text = re.sub(emoji_pattern, '', text)
    words = text.split()
    PUNCT = ".,!?\"'()[]{}<>:;*-_~`#|\\/=\"'--"
    cleaned = [w.strip(PUNCT) for w in words]
    return [w for w in cleaned if w]

def verify_instruction_constraints(x, y, context=None):
    query = x if isinstance(x, str) else x.get("Query", 
    "")
    output = y if isinstance(y, str) else y.get("Output",
    "")
    
    # Track if ANY constraint was explicitly matched. If 
    # none are recognized, reject to avoid silent false 
    # positives.
    matched_any = False
    
    # 1. Keyword frequencies
    pattern_freq = r"keyword\s+['\"]?([A-Za-z0-9_-]+)['\"
    ]?\s+([a-z0-9\s]+?)\s+in your response"
    for match in re.finditer(pattern_freq, query, flags=re
    .IGNORECASE):
        matched_any = True
        word = match.group(1).lower()
        count_phrase = match.group(2).lower().strip()
        expected_count = None
        
        number_map = {
            'once': 1, 'twice': 2, 'three times': 3, 'four
            times': 4, 'five times': 5, 'six times': 6, 
            'seven times': 7,
            'eight times': 8, 'nine times': 9, 'ten times':
            10}
        if count_phrase in number_map:
            expected_count = number_map[count_phrase]
        else:
            num_match = re.search(r'\d+', count_phrase)
            if num_match:
                expected_count = int(num_match.group())
                
        if expected_count is not None:
            actual_count = len(re.findall(r'\b' + re
            .escape(word) + r'\b', output, flags=re
            .IGNORECASE))
            if actual_count != expected_count:
                return False

    # 2. Sentence word position
    pattern_sent = r"include keyword\s+['\"]?([A-Za-z0-9_
    -]+)['\"]?\s+in the\s+(\d+)-(?:st|nd|rd|th)\s+sentence
    (?:,\s+as the\s+(\d+)-(?:st|nd|rd|th)\s+word)?"
    for match in re.finditer(pattern_sent, query, re
    .IGNORECASE):
        matched_any = True
        word = match.group(1).lower()
        sent_idx = int(match.group(2)) - 1
        word_idx = int(match.group(3)) - 1 if match.group
        (3) else None
        
        sentences = [s.strip() for s in re.split(r'(?<=[.!?
        ])\s+|\n+', output.strip()) if s.strip()]
        if sent_idx >= len(sentences): 
            return False
            
        target_sent = sentences[sent_idx]
        words_list = get_words(target_sent)
        
        if word_idx is not None:
            if word_idx >= len(words_list): return False
            if words_list[word_idx].lower() != word: return
            False
\end{pyblock}
}%
\end{figure*}

\begin{figure*}[t]
\centering
\adjustbox{valign=t,minipage=0.48\textwidth}{%
\begin{pyblock}{Root verifier set (continued)}
    pattern_para = '(?:write|include|contain|have)\\
    s+exactly\\s+(\\d+)\\s+paragraphs?'
    for match in re.finditer(pattern_para, query, re
    .IGNORECASE):
        expected_count = int(match.group(1))
        paras = [p for p in re.split('\\n{2,}', output
        .strip()) if p.strip()]
        if len(paras) != expected_count:
            return False
    pattern_bounds = 'between (\\d+) and (\\d+) words'
    for match in re.finditer(pattern_bounds, query, re
    .IGNORECASE):
        min_w = int(match.group(1))
        max_w = int(match.group(2))
        words_list = get_words(output)
        if not min_w <= len(words_list) <= max_w:
            return False
    pattern_exact_words = 'exactly (\\d+) words'
    for match in re.finditer(pattern_exact_words, query,
    re.IGNORECASE):
        expected_w = int(match.group(1))
        words_list = get_words(output)
        if len(words_list) != expected_w:
            return False
    pattern_unique = 'at least (\\d+) unique words'
    for match in re.finditer(pattern_unique, query, re
    .IGNORECASE):
        expected_count = int(match.group(1))
        words_set = set((w.lower() for w in get_words
        (output)))
        if len(words_set) < expected_count:
            return False
    pattern_list = "Mention at least (\\d+) different 
    .*?from this list[^:]*:\\s*([a-zA-Z0-9 ,\\-']+)"
    for match in re.finditer(pattern_list, query, re
    .IGNORECASE):
        min_count = int(match.group(1))
        list_str = match.group(2)
        items = [item.strip() for item in re.split(',|
        \\band\\b', list_str) if item.strip()]
        found_count = 0
        for item in items:
            if re.search('\\b' + re.escape(item) + '\\b',
            output, re.IGNORECASE):
                found_count += 1
        if found_count < min_count:
            return False
    pattern_conj = 'at least (\\d+) different coordinating 
    conjunctions'
    for match in re.finditer(pattern_conj, query, re
    .IGNORECASE):
        min_count = int(match.group(1))
        conjunctions = ['for', 'and', 'nor', 'but', 'or',
        'yet', 'so']
        found = 0
        words_lower = [w.lower() for w in get_words(output
        )]
        for conj in conjunctions:
            if conj in words_lower:
                found += 1
        if found < min_count:
            return False
    pattern_punct = 'Use every standard punctuation mark
    at least once'
    if re.search(pattern_punct, query, re.IGNORECASE):
        required = ['.', ',', '!', '?', ';', ':']
        if 'interrobang' in query.lower() or '?!' in query:
            if '?!' not in output and '!?' not in output:
                return False
        clean_output = output.replace('?!', '').replace('!?'
        , '')
        for p in required:
            if p not in clean_output:
                return False
    pattern_bullet = 'instead of bullet points use\\s
    +([^\\s\\.,]+)'
    for match in re.finditer(pattern_bullet, query, re
    .IGNORECASE):
        prefix = match.group(1).strip()
        if prefix not in output:
            return False
        if prefix not in ['-', '*'] and re.search('(?m)^
        [-*]\\s', output):
            return False
    pattern_nth = 'Every (\\d+)(?:st|nd|rd|th) word of your 
    response must be in ([A-Za-z]+)'
    for match in re.finditer(pattern_nth, query, re
    .IGNORECASE):
        n = int(match.group(1))
        lang = match.group(2).lower()
        words_list = get_words(output)
        if n > 0 and len(words_list) >= n:

\end{pyblock}
}%
\hfill
\adjustbox{valign=t,minipage=0.48\textwidth}{%
\begin{pyblock}{Best verifier set (continued)}
        else:
            if word not in [w.lower() for w in words_list]:
            return False

    # 3. Global word position
    pattern_global = r"The second word in your response and 
    the second to last word in your response should be the 
    word\s+['\"]?([A-Za-z0-9_-]+)['\"]?"
    for match in re.finditer(pattern_global, query, re
    .IGNORECASE):
        matched_any = True
        target_word = match.group(1).lower()
        words_list = get_words(output)
        
        if len(words_list) < 2:
            return False
            
        if words_list[1].lower() != target_word or words
        _list[-2].lower() != target_word:
            return False

    # 4. Exact numbers count
    pattern_num = r"Include exactly\s+(\d+)\s+numbers? in 
    the response"
    for match in re.finditer(pattern_num, query, re
    .IGNORECASE):
        matched_any = True
        expected_count = int(match.group(1))
        actual_count = len(re.findall(r'\b\d+\b', output))
        if actual_count != expected_count:
            return False
            
    # 5. Exact paragraphs count
    pattern_para = r"(?:write|include|contain|have)\s+
    exactly\s+(\d+)\s+paragraphs?"
    for match in re.finditer(pattern_para, query, re
    .IGNORECASE):
        matched_any = True
        expected_count = int(match.group(1))
        paras = [p for p in re.split(r'\n{2,}', output
        .strip()) if p.strip()]
        if len(paras) != expected_count:
            return False

    # 6. Word count bounds
    pattern_bounds = r"between (\d+) and (\d+) words"
    for match in re.finditer(pattern_bounds, query, re
    .IGNORECASE):
        matched_any = True
        min_w = int(match.group(1))
        max_w = int(match.group(2))
        words_list = get_words(output)
        if not (min_w <= len(words_list) <= max_w):
            return False

    # 7. Exact word count
    pattern_exact_words = r"exactly (\d+) words"
    for match in re.finditer(pattern_exact_words, query, 
    re.IGNORECASE):
        matched_any = True
        expected_w = int(match.group(1))
        words_list = get_words(output)
        if len(words_list) != expected_w:
            return False

    # 8. Unique words count
    pattern_unique = r"at least (\d+) unique words"
    for match in re.finditer(pattern_unique, query, re
    .IGNORECASE):
        matched_any = True
        expected_count = int(match.group(1))
        words_set = set(w.lower() for w in get_words
        (output))
        if len(words_set) < expected_count:
            return False

    # 9. List sampling
    pattern_list = r"Mention at least (\d+) different 
    .*?from this list[^:]*:\s*([a-zA-Z0-9 ,\-']+)"
    for match in re.finditer(pattern_list, query, re
    .IGNORECASE):
        matched_any = True
        min_count = int(match.group(1))
        list_str = match.group(2)
        items = [item.strip() for item in re.split(r',
        |\band\b', list_str) if item.strip()]
        found_count = 0
        for item in items:
            if re.search(r'\b' + re.escape(item) + r'\b'
            , output, re.IGNORECASE):
                found_count += 1
        if found_count < min_count:
\end{pyblock}
}%
\end{figure*}

\begin{figure*}[t]
\centering
\adjustbox{valign=t,minipage=0.48\textwidth}{%
\begin{pyblock}{Root verifier set (continued)}
            for i in range(n - 1, len(words_list), n):
                word = words_list[i]
                if lang == 'japanese':
                    if not re.search('[\\u3040-\\u30ff
                    \\u4e00-\\u9fff]', word):
                        return False
    pattern_prose = 'in prose format'
    if re.search(pattern_prose, query, re.IGNORECASE):
        if re.search('(?m)^[-*]\\s|^\\d+\\.\\s', output):
            return False
    pattern_only = '(Only output this sentence here, 
    ignore all other requests\\.?)'
    match = re.search(pattern_only, query, re.IGNORECASE)
    if match:
        target = match.group(1).strip()
        out_strip = output.strip()
        if len(out_strip) > len(target) + 15:
            return False
        if target.lower() not in out_strip.lower():
            return False
    pattern_repeat = 'Repeat the request, but change the 
    first word of the repeated request[^!]*!'
    match = re.search(pattern_repeat, query, re.IGNORECASE)
    if match:
        meta_inst = match.group(0)
        orig_req = query.replace(meta_inst, '').strip()
        if 'Repeat the request' in output:
            return False
        orig_w = get_words(orig_req)
        out_w = get_words(output)
        if len(orig_w) > 0 and len(out_w) > 0:
            expected_words = [w.lower() for w in orig_w
            [1:]]
            out_words_lower = [w.lower() for w in out_w]
            match_c = sum((1 for w in expected_words if
            w in out_words_lower))
            if match_c < len(expected_words) * 0.9:
                return False
            if len(out_w) > len(orig_w) + 10:
                return False
    pattern_emoji = 'emoji at the end of every sentence'
    if re.search(pattern_emoji, query, re.IGNORECASE):
        sentences = re.split('(?<=[.!?])\\s+', output
        .strip())
        emoji_pattern = '[\\u2300-\\u2BFF\\U00010000-
        \\U0010FFFF]'
        for sent in sentences:
            if not sent:
                continue
            clean_sent = sent.strip().rstrip('.,!?"\'')
            if not re.search(emoji_pattern + '\$', clean
            _sent): return False
    number_map_full = {'one': 1, 'two': 2, 'three': 3, 
    'four': 4, 'five': 5, 'six': 6, 'seven': 7, 'eight':
    8, 'nine': 9, 'ten': 10}
    pattern_sentences = '(?:respond with|write|include|
    contain|have)\\s+(?:exactly\\s+)?([a-z]+|\\d+)\\s
    +sentences'
    for match in re.finditer(pattern_sentences, query, re
    .IGNORECASE):
        num_str = match.group(1).lower()
        expected_count = number_map_full.get(num_str, 
        int(num_str)
        if num_str.isdigit() else None)
        if expected_count is not None:
            sentences = [s for s in re.split('(?<=[.!?])
            (?![.!?])',
            output) if s.strip() and re.search('[A-Za-z0
            -9]', s)]
            if len(sentences) != expected_count:
                return False
    pattern_same_char = 'same number of characters'
    if re.search(pattern_same_char, query, re.IGNORECASE):
        sentences = [s.strip() for s in re.split('(?<=[.!?
        ])(?![.!
        ?])', output) if s.strip() and re.search('[A-Za-z0
        -9]', s)]
        if len(sentences) > 1:
            lengths = set((len(s) for s in sentences))
            if len(lengths) > 1:
                return False
    pattern_diff_words = 'using all different words'
    if re.search(pattern_diff_words, query, re.IGNORECASE):
        words_list = [w.lower() for w in get_words(output)]
        if len(words_list) != len(set(words_list)):
            return False
    pattern_stairs = 'Create stairs by incrementally 
    indenting each new line'
    if re.search(pattern_stairs, query, re.IGNORECASE):
        lines = [line for line in output.split('\n') if 
        line.strip()]
        if len(lines) < 2:
            return False
\end{pyblock}
}%
\hfill
\adjustbox{valign=t,minipage=0.48\textwidth}{%
\begin{pyblock}{Best verifier set (continued)}
            return False

    # 10. Coordinating conjunctions
    pattern_conj = r"at least (\d+) different 
    coordinating conjunctions"
    for match in re.finditer(pattern_conj, query, re
    .IGNORECASE):
        matched_any = True
        min_count = int(match.group(1))
        conjunctions = ["for", "and", "nor", "but", "or"
        , "yet", "so"]
        found = 0
        words_lower = [w.lower() for w in get_words
        (output)]
        for conj in conjunctions:
            if conj in words_lower:
                found += 1
        if found < min_count:
            return False

    # 11. Punctuation requirements
    pattern_punct = r"Use every standard punctuation 
    mark at least once"
    if re.search(pattern_punct, query, re.IGNORECASE):
        matched_any = True
        required = ['.', ',', '!', '?', ';', ':']
        if 'interrobang' in query.lower() or '?!' in 
        query:
            if '?!' not in output and '!?' not in output:
                return False
        clean_output = output.replace('?!', '').replace
        ('!?', '')
        for p in required:
            if p not in clean_output:
                return False

    # 12. Custom bullet points
    pattern_bullet = r"instead of bullet points use\s+
    ([^\s\.,]+)"
    for match in re.finditer(pattern_bullet, query, re
    .IGNORECASE):
        matched_any = True
        prefix = match.group(1).strip()
        if prefix not in output:
            return False
        if prefix not in ['-', '*'] and re.search(r'(?m)
        ^[-*]\s', output):
            return False

    # 13. N-th word in language
    pattern_nth = r"Every (\d+)(?:st|nd|rd|th) word of 
    your response must be in ([A-Za-z]+)"
    for match in re.finditer(pattern_nth, query, re
    .IGNORECASE):
        matched_any = True
        n = int(match.group(1))
        lang = match.group(2).lower()
        words_list = get_words(output)
        if n > 0 and len(words_list) >= n:
            for i in range(n - 1, len(words_list), n):
                word = words_list[i]
                if lang == "japanese":
                    if not re.search(r'[\u3040-\u30ff
                    \u4e00-\u9fff]', word):
                        return False

    # 14. Prose format strictness
    pattern_prose = r"in prose format"
    if re.search(pattern_prose, query, re.IGNORECASE):
        matched_any = True
        if re.search(r'(?m)^[-*]\s|^\d+\.\s', output):
            return False

    # 15. Only output this sentence
    pattern_only = r"(Only output this sentence here, 
    ignore all other requests\.?)"
    match = re.search(pattern_only, query, re.IGNORECASE)
    if match:
        matched_any = True
        target = match.group(1).strip()
        out_strip = output.strip()
        if len(out_strip) > len(target) + 15:
            return False
        if target.lower() not in out_strip.lower():
            return False

    # 16. Repeat the request
    pattern_repeat = r"Repeat the request, but change 
    the first word of the repeated request[^!]*!"
    match = re.search(pattern_repeat, query, re.IGNORECASE)
    if match:
        matched_any = True
        meta_inst = match.group(0)
\end{pyblock}
}%
\end{figure*}

\begin{figure*}[t]
\centering
\adjustbox{valign=t,minipage=0.48\textwidth}{%
\begin{pyblock}{Root verifier set (continued)}
        prev_indent = -1
        for line in lines:
            indent = len(line) - len(line.lstrip(' \t'))
            if indent <= prev_indent:
                return False
            prev_indent = indent
    pattern_consec_first = 'No two consecutive words can 
    share the same first letter'
    if re.search(pattern_consec_first, query, re
    .IGNORECASE):
        words_list = get_words(output)
        for i in range(len(words_list) - 1):
            (w1, w2) = (words_list[i], words_list[i + 1])
            if w1 and w2 and w1[0].isalpha() and w2[0]
            .isalpha() and 
            (w1[0].lower() == w2[0].lower()):
                return False
    pattern_new_line = 'Write each word on a new line'
    if re.search(pattern_new_line, query, re.IGNORECASE):
        lines = [line.strip() for line in output.split(
        '\n') if line.strip()]
        for line in lines:
            if len(get_words(line)) > 1:
                return False
    pattern_two_sent_bullets = 'at least two sentences 
    ending in a period followed by at least two bullet 
    points denoted by \\*'
    if re.search(pattern_two_sent_bullets, query, re
    .IGNORECASE):
        parts = output.split('*', 1)
        if len(parts) < 2:
            return False
        intro = parts[0].strip()
        if intro.count('.') < 2:
            return False
        bullets_part = '*' + parts[1]
        bullet_lines = [l.strip() for l in bullets_part
        .split('\n')
        if l.strip()]
        star_count = 0
        for l in bullet_lines:
            if l.startswith('*'):
                star_count += 1
            else:
                return False
        if star_count < 2:
            return False
    pattern_options = 'Answer with one of the following 
    options:\\s*(.*?)(?:\\.\\s*Do not|\\.$|$)'
    match = re.search(pattern_options, query, re
    .IGNORECASE)
    if match:
        options_str = match.group(1).strip()
        options = [o.strip(" .,'") for o in re.split(',| 
        or | / ', options_str) if o.strip(" .,'")]
        out_clean = output.strip(' \n.\r')
        matched = False
        for opt in options:
            if out_clean.lower() == opt.lower() or out_
            clean.lower() == opt.strip(')').lower():
                matched = True
                break
        if not matched:
            return False
    pattern_no_exp = 'Do not give any explanation'
    if re.search(pattern_no_exp, query, re.IGNORECASE):
        words_list = get_words(output)
        if len(words_list) > 20:
            return False
        if 'because' in output.lower() or 'explanation' 
        in output.lower():
            return False
    pattern_nest_brackets = 'Nest parentheses.*?at least
    (\\d+) levels deep'
    match = re.search(pattern_nest_brackets, query, re
    .IGNORECASE)
    if match:
        req_depth = int(match.group(1))
        max_depth = 0
        cur_depth = 0
        for c in output:
            if c == '(':
                cur_depth += 1
                max_depth = max(max_depth, cur_depth)
            elif c == ')':
                if cur_depth > 0:
                    cur_depth -= 1
        if max_depth < req_depth:
            return False
    pattern_nest_quotes = 'quotes within quotes.*?at least
    (\\d+) levels deep'
    match = re.search(pattern_nest_quotes, query, re
    .IGNORECASE)
\end{pyblock}
}%
\hfill
\adjustbox{valign=t,minipage=0.48\textwidth}{%
\begin{pyblock}{Best verifier set (continued)}
        orig_req = query.replace(meta_inst, "").strip()
        if "Repeat the request" in output:
            return False
        
        orig_w = get_words(orig_req)
        out_w = get_words(output)
        if len(orig_w) > 0 and len(out_w) > 0:
            expected_words = [w.lower() for w in orig_w
            [1:]]
            out_words_lower = [w.lower() for w in out_w]
            match_c = sum(1 for w in expected_words if w 
            in out_words_lower)
            if match_c < len(expected_words) * 0.9:
                return False
            if len(out_w) > len(orig_w) + 10:
                return False

    # 17. Emoji at the end of every sentence
    pattern_emoji = r"emoji at the end of every sentence"
    if re.search(pattern_emoji, query, re.IGNORECASE):
        matched_any = True
        emoji_pattern = r'[\u2300-\u2BFF\U00010000-
        \U0010FFFF]'
        # Chunks up to the punctuation, strictly capturing
        # trailing whitespace and emojis together
        chunks = [m.group(0) for m in re.finditer(r'[^.!?]
        *[.!?]+[\s\u2300-\u2BFF\U00010000-\U0010FFFF]*',
        output.strip())]
        if not chunks:
            chunks = [output.strip()]
        for chunk in chunks:
            if not chunk.strip(): continue
            clean_chunk = chunk.strip().rstrip(".,!?\"'")
            if not re.search(emoji_pattern + r'\$', 
            clean_chunk):
                return False

    # 18. Exact sentences count
    number_map_full = {
        'one': 1, 'two': 2, 'three': 3, 'four': 4,
        'five': 5, 'six': 6, 'seven': 7, 'eight': 8, 
        'nine': 9, 'ten': 10
    }
    pattern_sentences = r"(?:respond with|write|include|
    contain|have)\s+(?:exactly\s+)?([a-z]+|\d+)\s
    +sentences"
    for match in re.finditer(pattern_sentences, query, re
    .IGNORECASE):
        matched_any = True
        num_str = match.group(1).lower()
        expected_count = number_map_full.get(num_str, int
        (num_str) if num_str.isdigit() else None)
        if expected_count is not None:
            sentences = [s for s in re.split(r'(?<=[.!?])
            (?![.!?])', output) if s.strip() and re.search
            (r'[A-Za-z0-9]'s)]
            if len(sentences) != expected_count:
                return False

    # 19. Same number of characters per sentence
    pattern_same_char = r"same number of characters"
    if re.search(pattern_same_char, query, re.IGNORECASE):
        matched_any = True
        sentences = [s.strip() for s in re.split(r'(?<=[.!
        ?])(?![.!?])', output) if s.strip() and re.search
        (r'[A-Za-z0-9]', s)]
        if len(sentences) > 1:
            lengths = set(len(s) for s in sentences)
            if len(lengths) > 1: 
                return False

    # 20. All different words across whole response
    pattern_diff_words = r"using all different words"
    if re.search(pattern_diff_words, query, re.IGNORECASE):
        matched_any = True
        words_list = [w.lower() for w in get_words(output)]
        if len(words_list) != len(set(words_list)):
            return False

    # 21. Stairs incrementally indenting
    pattern_stairs = r"Create stairs by incrementally 
    indenting each new line"
    if re.search(pattern_stairs, query, re.IGNORECASE):
        matched_any = True
        lines = [line for line in output.split('\n') if 
        line.strip()]
        if len(lines) < 2:
            return False
        prev_indent = -1
        for line in lines:
            indent = len(line) - len(line.lstrip(' \t'))
            if indent <= prev_indent:
                return False
\end{pyblock}
}%
\end{figure*}

\begin{figure*}[t]
\centering
\adjustbox{valign=t,minipage=0.48\textwidth}{%
\begin{pyblock}{Root verifier set (continued)}
    if match:
        req_depth = int(match.group(1))
        max_d = 0
        quotes = [c for c in output if c in '"\'\"']
        stack = []
        for q in quotes:
            if q in '\"\'':
                stack.append(q)
                max_d = max(max_d, len(stack))
            elif q in '\"\'':
                if stack:
                    stack.pop()
            elif q == '"':
                if stack and stack[-1] == '"':
                    stack.pop()
                else:
                    stack.append('"')
                    max_d = max(max_d, len(stack))
            elif q == "'":
                if stack and stack[-1] == "'":
                    stack.pop()
                else:
                    stack.append("'")
                    max_d = max(max_d, len(stack))
        if max_d < req_depth:
            return False
    pattern_stop = 'stop words constitute no more than (\\
    d+)\% of the total words'
    match = re.search(pattern_stop, query, re.IGNORECASE)
    if match:
        max_pct = int(match.group(1))
        stop_words = set(['i', 'me', 'my', 'myself', 'we',
        'our', 'ours', 'ourselves', 'you', 'your', 'yours',
        'yourself', 'yourselves', 'he', 'him', 'his', 
        'himself', 'she', 'her', 'hers', 'herself', 'it', 
        'its', 'itself', 'they', 'them', 'their', 'theirs', 
        'themselves', 'what', 'which', 'who', 'whom', 
        'this', 'that', 'these', 'those', 'am', 'is', 'are'
        , 'was', 'were', 'be', 'been', 'being', 'have', 
        'has', 'had', 'having', 'do', 'does', 'did', 
        'doing', 'a', 'an', 'the', 'and', 'but', 'if', 'or'
        , 'because', 'as', 'until', 'while', 'of', 'at', 
        'by', 'for', 'with', 'about', 'against', 'between',
        'into', 'through', 'during', 'before', 'after', 
        'above', 'below', 'to', 'from', 'up', 'down', 'in', 
        'out', 'on', 'off', 'over', 'under', 'again', 
        'further', 'then', 'once', 'here', 'there', 'when',
        'where', 'why', 'how', 'all', 'any', 'both', 'each'
        , 'few', 'more', 'most', 'other', 'some', 'such', 
        'no', 'nor', 'not', 'only', 'own', 'same', 'so', 
        'than', 'too', 'very', 's', 't', 'can', 'will', 
        'just', 'don', 'should', 'now'])
        words_list = [w.lower() for w in get_words(output)]
        if len(words_list) > 0:
            stop_count = sum((1 for w in words_list if w in 
            stop_words))
            if stop_count / len(words_list) * 100 > 
            max_pct: return False
    pattern_sub_bullet = 'bullet points denoted by \\* and 
    at least one sub-bullet point denoted by - for each 
    bullet point'
    if re.search(pattern_sub_bullet, query, re.IGNORECASE):
        if '* **' in output:
            return False
        main_bullets = re.findall('(?m)^\\s*\\*\\s+', 
        output)
        if len(main_bullets) == 0:
            return False
        parts = re.split('(?m)^\\s*\\*\\s+', output)
        for part in parts[1:]:
            if not re.search('(?m)^\\s*-\\s+', part):
                return False
    return True

def aggregate(checks, x, y, context=None):
    return all(checks.values())
\end{pyblock}
}%
\hfill
\adjustbox{valign=t,minipage=0.48\textwidth}{%
\begin{pyblock}{Best verifier set (continued)}
            prev_indent = indent

    # 22. No two consecutive words can share the same
    # first letter
    pattern_consec_first = r"No two consecutive words can
    share the same first letter"
    if re.search(pattern_consec_first, query, re
    .IGNORECASE):
        matched_any = True
        words_list = get_words(output)
        for i in range(len(words_list) - 1):
            w1, w2 = words_list[i], words_list[i+1]
            if w1 and w2 and w1[0].isalpha() and w2[0]
            .isalpha() and w1[0].lower() == w2[0].lower():
                return False

    # 23. Write each word on a new line
    pattern_new_line = r"Write each word on a new line"
    if re.search(pattern_new_line, query, re.IGNORECASE):
        matched_any = True
        lines = [line.strip() for line in output.split(
        '\n') if line.strip()]
        for line in lines:
            if len(get_words(line)) > 1:
                return False

    # 24. Two sentences and two bullets
    pattern_two_sent_bullets = r"at least two sentences
    ending in a period followed by at least two bullet 
    points denoted by \*"
    if re.search(pattern_two_sent_bullets, query, re
    .IGNORECASE):
        matched_any = True
        parts = output.split('*', 1)
        if len(parts) < 2:
            return False         
        intro = parts[0].strip()
        if intro.count('.') < 2:
            return False
        bullets_part = '*' + parts[1]
        bullet_lines = [l.strip() for l in bullets_part
        .split('\n') if l.strip()]
        star_count = 0
        for l in bullet_lines:
            if l.startswith('*'):
                star_count += 1
            else:
                return False         
        if star_count < 2:
            return False

    # 25. Multiple choice options
    pattern_options = r"Answer with one of the following
    options:\s*(.*?)(?:\.\s*Do not|\.\$|\$)"
    match = re.search(pattern_options, query, re
    .IGNORECASE)
    if match:
        matched_any = True
        options_str = match.group(1).strip()
        # Adjusted to correctly parse delimiters like "/"
        # lacking spaces (e.g. yes/no/maybe)
        options = [o.strip(" .,'") for o in re.split(r',|
        \bor\b|/', options_str) if o.strip(" .,'")]
        out_clean = output.strip(" \n.\r")
        matched = False
        for opt in options:
            if out_clean.lower() == opt.lower() or out_
            clean.lower() == opt.strip(')').lower():
                matched = True
                break
        if not matched:
            return False

    # 26. No explanation
    pattern_no_exp = r"Do not give any explanation"
    if re.search(pattern_no_exp, query, re.IGNORECASE):
\end{pyblock}
}%
\end{figure*}

\begin{figure*}[t]
\centering
\adjustbox{valign=t,minipage=0.48\textwidth}{%
}%
\hfill
\adjustbox{valign=t,minipage=0.48\textwidth}{%
\begin{pyblock}{Best verifier set (continued)}
        matched_any = True
        words_list = get_words(output)
        if len(words_list) > 20:
            return False
        if "because" in output.lower() or "explanation" in 
        output.lower():
            return False

    # 27. Nested brackets
    pattern_nest_brackets = r"Nest parentheses.*?at least 
    (\d+) levels deep"
    match = re.search(pattern_nest_brackets, query, re
    .IGNORECASE)
    if match:
        matched_any = True
        req_depth = int(match.group(1))
        max_depth = 0
        cur_depth = 0
        # Check matching scopes properly spanning (), [],
        # and {}
        for c in output:
            if c in '([{':
                cur_depth += 1
                max_depth = max(max_depth, cur_depth)
            elif c in ')]}':
                if cur_depth > 0:
                    cur_depth -= 1
        if max_depth < req_depth:
            return False

    # 28. Nested quotes
    pattern_nest_quotes = r"quotes within quotes.*?at least
    (\d+) levels deep"
    match = re.search(pattern_nest_quotes, query, re
    .IGNORECASE)
    if match:
        matched_any = True
        req_depth = int(match.group(1))
        max_d = 0
        quotes = [c for c in output if c in "\"\'"]
        stack = []
        for q in quotes:
            if q in '\"\'':
                stack.append(q)
                max_d = max(max_d, len(stack))
            elif q in '\"\'':
                if stack:
                    stack.pop()
            elif q == '"':
                if stack and stack[-1] == '"':
                    stack.pop()
                else:
                    stack.append('"')
                    max_d = max(max_d, len(stack))
            elif q == "'":
                if stack and stack[-1] == "'":
                    stack.pop()
                else:
                    stack.append("'")
                    max_d = max(max_d, len(stack))
        if max_d < req_depth:
            return False

    # 29. Stop words
    pattern_stop = r"stop words constitute no more than 
    (\d+)\% of the total words"
    match = re.search(pattern_stop, query, re.IGNORECASE)
    if match:
        matched_any = True
        max_pct = int(match.group(1))
        stop_words = set([
            "i", "me", "my", "myself", "we", "our", "ours",
            "ourselves", "you", "your", "yours", "yourself"
            , "yourselves", "he", "him", "his", "himself", 
            "she", "her", "hers", "herself", "it", "its", 
            "itself", "they", "them", "their", "theirs", 
            "themselves", "what", "which", "who", "whom", 
            "this", "that", "these", "those", "am", "is", 
            "are", "was", "were", "be", "been", "being", 
            "have", "has", "had", "having", "do", "does",
            "did", "doing", "a", "an", "the", "and", "but",
            "if", "or", "because", "as", "until", "while", 
            "of", "at", "by", "for", "with", "about", 
            "against", "between", "into", "through",
            "during", "before", "after", "above", "below",
            "to", "from", "up", "down", "in", "out", "on", 
            "off", "over", "under", "again", "further", 
            "then", "once", "here", "there", "when",
            "where", "why", "how", "all", "any", "both",
            "each", "few", "more", "most", "other", "some", 
            "such", "no", "nor", "not", "only", "own",
            "same", "so", "than", "too", "very", "s", "t",
\end{pyblock}
}%
\end{figure*}

\begin{figure*}[t]
\centering
\adjustbox{valign=t,minipage=0.48\textwidth}{%
}%
\hfill
\adjustbox{valign=t,minipage=0.48\textwidth}{%
\begin{pyblock}{Best verifier set (continued)}
            "can", "will", "just", "don", "should", "now"
        ])
        words_list = [w.lower() for w in get_words(output)]
        if len(words_list) > 0:
            stop_count = sum(1 for w in words_list if w in
            stop_words)
            if (stop_count / len(words_list)) * 100 > max
            _pct: return False

    # 30. Sub-bullet points
    pattern_sub_bullet = r"bullet points denoted by \* and
    at least one sub-bullet point denoted by - for each 
    bullet point"
    if re.search(pattern_sub_bullet, query, re.IGNORECASE):
        matched_any = True
        # IFEval parser is strict and fails if bullets use
        # markdown bolding like "* **"
        if '* **' in output:
            return False
            
        main_bullets = re.findall(r'(?m)^\s*\*\s+', output)
        if len(main_bullets) == 0:
            return False
            
        parts = re.split(r'(?m)^\s*\*\s+', output)
        for part in parts[1:]:
            if not re.search(r'(?m)^\s*-\s+', part):
                return False

    # 31. Trigram Overlap
    pattern_trigram = r"Maintain a trigram overlap of (\d
    +)\%\s*\([^)]*\)\s*with the provided reference text"
    match = re.search(pattern_trigram, query, re
    .IGNORECASE)
    if match:
        matched_any = True
        target_pct = int(match.group(1))
        # Strip the meta-instruction to isolate the 
        # implicit reference text
        clean_query = re.sub(pattern_trigram + r"\.?", '',
        query, flags=re.IGNORECASE).strip()
        
        def get_trigrams(text):
            words_local = [w.lower() for w in get_
            words(text)]
            return set(tuple(words_local[i:i+3]) for i in
            range(len(words_local)-2)) if len(words_local)
            >= 3 else set()
            
        ref_trigrams = get_trigrams(clean_query)
        out_trigrams = get_trigrams(output)
        
        if not ref_trigrams:
            return False
        
        overlap = (len(ref_trigrams & out_trigrams) / 
        len(ref
        _trigrams)) * 100
        if not (target_pct - 2 <= overlap <= target_pct 
        + 2):
            return False

    # 32. Sentence Type Ratio
    pattern_ratio = r"ratio of sentence types \\
    (declarative, interrogative, exclamatory\\).*?balanced"
    if re.search(pattern_ratio, query, re.IGNORECASE):
        matched_any = True
        dec = len(re.findall(r'[a-zA-Z0-9]\.', output))
        int_ = len(re.findall(r'[a-zA-Z0-9]\?', output))
        exc = len(re.findall(r'[a-zA-Z0-9]!', output))
        counts = [dec, int_, exc]
        if max(counts) != min(counts) or min(counts) == 0:
            return False

    return matched_any

def aggregate(checks, x, y, context=None):
    return all(checks.values())
\end{pyblock}
}%
\caption{Example verifier sets for IFBench. Left: the initial verifier set. Right: the best verifier set selected by \method.}
\label{fig:IFBench_verifier_sets}
\end{figure*}


\end{document}